\documentclass{article}

\usepackage{PRIMEarxiv}

\usepackage[utf8]{inputenc}
\usepackage[T1]{fontenc}
\usepackage[hidelinks]{hyperref}
\usepackage{url}
\usepackage{booktabs}
\usepackage{amsfonts}
\usepackage{nicefrac}
\usepackage{microtype}
\usepackage{lipsum}
\usepackage{fancyhdr}
\usepackage{graphicx}

\usepackage{array}
\usepackage{amsmath}
\usepackage{enumitem}
\newlist{rqs}{enumerate}{1}
\setlist[rqs]{label*=\textbf{RQ\arabic*}}
\newcommand\bulletparagraph[1]{\quad\textbullet\ \textbf{#1}}

\title{Large Language Models for Forecasting and Anomaly Detection: A Systematic Literature Review}

\author{
  Jing Su \\
  AT\&T Center for Virtualization \\
  Southern Methodist University \\
  Dallas, TX, USA\\
  \texttt{suj@smu.edu} \\
  \And
  Chufeng Jiang \\
  Department of Computer Science \\
  The University of Texas at Austin \\
  Austin, TX, USA\\
  \texttt{chufeng.jiang@utexas.edu} \\
  \And
  Xin Jin \\
  Department of Computer Science and Engineering \\
  Ohio State University \\
  Columbus, OH, USA\\
  \texttt{jin.967@osu.edu} \\
  \And
  Yuxin Qiao \\
  Department of Information System \\
  Universidad Internacional Isabel I de Castilla \\
  Burgos, Spain \\
  \texttt{qiaoyuxin46@icloud.com} \\
  \And
  Tingsong Xiao \\
  Department of Computer \& Information Science \& Engineering \\
  University of Florida \\
  Gainesville, FL, USA \\
  \texttt{xiaotingsong@ufl.edu} \\
  \And
  Hongda Ma \\
  Department of Computer Science \\
  The University of Texas at Austin \\
  Austin, TX, USA \\
  \texttt{hongda.ma@utexas.edu} \\
  \And
  Rong Wei \\
  Academy for Advanced Interdisciplinary Studies \\
  Peking University \\
  Beijing, China \\
  \texttt{wei\_rong@pku.edu.cn} \\
  \And
  Zhi Jing \\
  School of Computer Science \\
  Carnegie Mellon University \\
  Pittsburgh, PA, USA \\
  \texttt{zjing2@cs.cmu.edu} \\
  \And
  Jiajun Xu \\
  Department of Electrical and Computer Engineering\\
  University of Southern California \\
  Los Angeles, CA, USA \\
  \texttt{jiajunx@usc.edu} \\
  \And
  Junhong Lin \\
  Electrical Engineering \& Computer Science Department \\
  Massachusetts Institute of Technology \\
  Cambridge, MA, USA \\
  \texttt{junhong@mit.edu} \\
}

\begin{document}

\maketitle

\begin{abstract}
  This systematic literature review comprehensively examines the application of Large Language Models (LLMs) in forecasting and anomaly detection, highlighting the current state of research, inherent challenges, and prospective future directions. LLMs have demonstrated significant potential in parsing and analyzing extensive datasets to identify patterns, predict future events, and detect anomalous behavior across various domains. However, this review identifies several critical challenges that impede their broader adoption and effectiveness, including the reliance on vast historical datasets, issues with generalizability across different contexts, the phenomenon of model hallucinations, limitations within the models' knowledge boundaries, and the substantial computational resources required. Through detailed analysis, this review discusses potential solutions and strategies to overcome these obstacles, such as integrating multimodal data, advancements in learning methodologies, and emphasizing model explainability and computational efficiency. Moreover, this review outlines critical trends that are likely to shape the evolution of LLMs in these fields, including the push toward real-time processing, the importance of sustainable modeling practices, and the value of interdisciplinary collaboration. Conclusively, this review underscores the transformative impact LLMs could have on forecasting and anomaly detection while emphasizing the need for continuous innovation, ethical considerations, and practical solutions to realize their full potential.
\end{abstract}

\keywords{Large Language Models \and Pre-trained Foundation Models \and Time Series \and Forecasting \and Anomaly Detection}

\section{Introduction}\label{sec:introduction}

Language represents a rigorously structured communicative system characterized by its grammar and vocabulary. It serves as the principal medium through which humans articulate and convey meaning. This conception of language as a structured communicative tool is pivotal in the realm of computational linguistics, particularly in the development and evaluation of natural language processing (NLP) algorithms. A seminal aspect in this field is the Turing Test, proposed by Alan Turing in 1950 \cite{turing1950computing_machinery}, which evaluates a machine's ability to exhibit intelligent behavior equivalent to, or indistinguishable from, that of a human. In this context, the Turing Test primarily assesses the machine's capability to perform tasks involving language comprehension and generation, reflecting the intricate role of linguistic structure in the artificial replication of human-like communication.

Language model (LM) is a fundamental element employed in a multitude of NLP tasks, such as text generation, machine translation, and speech recognition~\cite{devlin2019bert_pre-training,openai2023gpt-4_technical}. These models are intricately designed to comprehend, generate, and manipulate human language. The training of language models involves large-scale corpora, enabling them to learn universal language representations. This training process is critical for the models to capture the semantics of words in varying contexts~\cite{sousa2019bert,jin2023understand, zhang2023identifying}. Notably, the fidelity of these representations is frequently contingent on the word frequency within the training corpus. Such dependency underscores the importance of a comprehensive and diverse corpus in training LMs, as it directly impacts their ability to reflect and understand the nuances of natural language accurately. The intricacy of language models and their reliance on corpus characteristics are vital considerations in advancing NLP technologies, which underscores the significance of human-like language comprehension and production in artificial intelligence systems.

The forefront of advancements in language model technology has been marked by the emergence of Large Language Models (LLMs). This evolution signifies a paradigm shift in the field of NLP and extends its impact to broader applications. LLMs leverage deep learning methodologies~\cite{lecun2015deep}, utilizing extensive datasets to perform complex tasks such as understanding, summarizing, generating, and predicting novel content. These models operate by processing an input text and iteratively predicting subsequent tokens or words. A distinguishing feature of LLMs is their vast parameter space, encompassing tens to hundreds of billion parameters, in stark contrast to their predecessors~\cite{sousa2019bert,openai2023gpt-4_technical}. In addition, they are trained on significantly larger datasets, ranging from several gigabytes to terabytes in size. This exponential increase in both computational capacity and training data volume has not only enhanced the performance of LLMs in conventional NLP tasks but also has expanded their applicability in areas such as contextual analysis and sentiment detection.
The advancements in LLMs reflect the ongoing pursuit of achieving and surpassing human-level proficiency in language understanding and generation.

Forecasting and anomaly detection represent pivotal components in the realm of data science, delivering essential insights across a multitude of domains ranging from network security to financial markets~\cite{liao2022self, cao2024tempo_prompt-based,chen2022bert-log_anomaly, nayak2023q, pourmajidi2021immutable, zhang2022m, wang2021teaching}. These techniques are integral in projecting forthcoming trends and pinpointing atypical patterns that diverge from normative expectations. Such capabilities are proactive in fostering preemptive strategies in a wide array of applications.

Forecasting uses historical data to make informed predictions about future occurrences or trends. It involves making assumptions about the situation being analyzed, selecting an appropriate data set, analyzing the data, and determining the forecast. Forecasting serves as a cornerstone for strategic planning and decision-making in diverse sectors, ranging from economics and finance to healthcare and environmental management, that empowers organizations and policymakers to anticipate changes, manage risks, and allocate resources efficiently~\cite{lim2021time, zhang2019smva, su2022optimal}. In the financial sector, for instance, accurate forecasting is essential for investment strategies, risk management, and market analysis~\cite{cao2019financial, liu2023financial}. It enables investors and financial analysts to predict market trends, assess the viability of investments, and mitigate potential risks. Similarly, in supply chain management, forecasting plays a pivotal role in inventory control, demand planning, and logistics optimization, thus ensuring operational efficiency and cost-effectiveness~\cite{pacella2021evaluation}. Moreover, in the realm of public policy and healthcare, forecasting is critical for preparing for future demands, whether it be anticipating economic shifts, public health needs, or environmental changes~\cite{zeger2006time}. Accurate predictions can guide policy formulation and resource allocation, thereby enhancing the effectiveness of public services and interventions.

Anomaly detection, also known as outlier detection, is an analytical process aimed at identifying data points, entities, or occurrences that exhibit significant deviations from the typical patterns or norms~\cite{zhang2019impact, dang2020time_series}.
This methodology plays a critical role in automated surveillance systems, particularly in identifying potentially detrimental outliers, thereby safeguarding data integrity and security~\cite{dang2021ts-bert_time}.
Its application is especially crucial in sectors such as finance~\cite{ahmed2016survey}, retail~\cite{kim2003design}, and cybersecurity~\cite{jin2022edge, sun2019understanding}.
In the financial industry, anomaly detection is instrumental in fraud detection and anti-money laundering efforts.
It enables financial institutions to quickly identify unusual transaction patterns that may indicate fraudulent activity, thereby protecting both the institution and its customers from financial loss~\cite{ahmed2016survey,wu2023enhanced}.
Similarly, in the retail sector, anomaly detection can highlight unusual purchasing patterns or inventory issues, assisting in loss prevention and optimizing supply chain management~\cite{kim2003design}. The field of cybersecurity significantly benefits from anomaly detection. It is used to identify unusual network traffic, access patterns, or system behavior that could signify a security breach or cyberattack~\cite{jin2022edge, sun2021having}. By detecting these anomalies early, organizations can rapidly respond to potential threats, mitigating the risk of data breaches and cyberattacks~\cite{yao2023survey,jin2023prometheus}.

\begin{figure}
    \centering
    \includegraphics[width=.6\columnwidth]{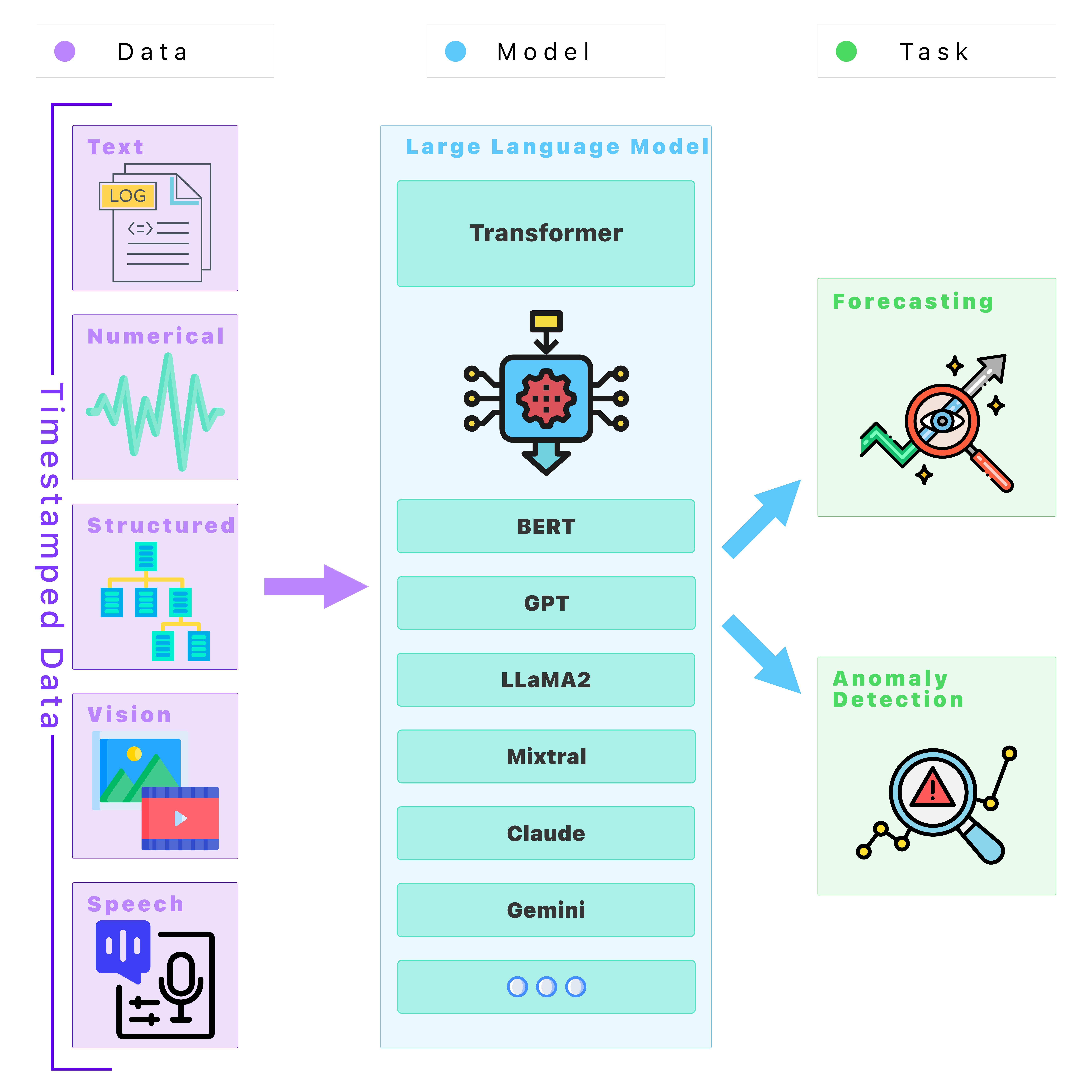}
    \caption{Overview of leveraging large language model for forecasting and anomaly detection tasks. The input data is often time series or timestamped data.}
    \label{fig:task-overview}
\end{figure}

Forecasting and anomaly detection are analytical processes inherently well-suited for time series or timestamped data due to the temporal nature of the information they seek to understand and leverage. Time series data, by definition, is a sequence of data points collected or recorded at time intervals, which often exhibits trends, seasonal variations, and cycles that forecasting techniques aim to capture and extrapolate into the future~\cite{dang2021ts-bert_time, dong2024prediction}. Timestamped data is particularly conducive to anomaly detection because it allows for the recognition of deviations from established temporal patterns. For instance, in cybersecurity, anomaly detection systems can identify unusual access patterns that may indicate a security breach~\cite{chen2022bert-log_anomaly}. In industrial settings, it might flag an unexpected drop or spike in sensor readings, potentially preventing equipment failure.\ \autoref{fig:task-overview} depicts an overview of leveraging a large language model for forecasting and anomaly detection tasks. The input data is often time series or timestamped data, encompassing a variety of formats such as text logs, numerical data, structured data, graphical input, and speech recordings. Current widely used LLMs such as BERT \cite{devlin2019bert_pre-training}, GPT \cite{radford2019language_models}, LLaMA2 \cite{touvron2023llama_2}, and Mixtral \cite{jiang2024mixtral_of} are transformer-based, which includes mechanisms such as input and output embeddings, multi-head attention, and feed-forward neural networks. Forecasting tasks involve predicting future data points based on learned patterns, while anomaly detection identifies outliers or unexpected events in the data stream.
\looseness=-1

In this study, we embark on a \textit{comprehensive exploration} of the integration and potential of LLMs in the realms of forecasting and anomaly detection, areas traditionally dominated by quantitative data analysis. The rapid evolution of LLMs in NLP presents an unprecedented opportunity to augment and possibly revolutionize these domains. This paper aims to bridge the gap between the advanced linguistic processing capabilities of LLMs and the predictive analytics involved in forecasting and detecting outliers. We delve into how the qualitative insights gleaned from LLMs can complement traditional quantitative approaches, thereby enriching the analytical depth and accuracy in various sectors, including finance, cybersecurity, and healthcare. Additionally, this survey addresses the challenges, ethical considerations, and future research trajectories at the intersection of LLMs with these critical data science applications. Our objective is to provide a holistic view that not only elucidates the current state of LLM applications in these fields but also stimulates interdisciplinary dialogue and research, navigating the complexities of modern data environments and paving the way for innovative solutions in predictive analytics.

\textbf{Contributions}. To recapitulate, we highlight the following contributions:

\begin{itemize}
    \item To the best of our knowledge, this is the first comprehensive systematic literature review (SLR) dedicated to the application of LLMs in the domains of forecasting and anomaly detection. Through this review, we have elucidated the distinctive influences of LLMs on both numerical and textual data within these specific tasks.
    \item This study compiled a set of guidelines that delineate the optimal utilization of LLMs for various tasks, contributing significantly to the field by providing a structured approach to employing these advanced models in practical scenarios.
    \item This literature review offers, as far as possible, a deep theoretical insight into the capabilities of LLMs, particularly in handling complex patterns and nuances in data that traditional models may overlook. This includes an exploration of the underlying mechanisms that enable LLMs to process and interpret both structured and unstructured data effectively.
    \item This work opens up the enlightenment of new paths for future works around forecasting and anomaly detection modeling.
\end{itemize}

\paragraph{Roadmap} The remainder of this paper is organized as follows. Section~\ref{sec:methodology} outlines the methodology employed in conducting the systematic literature review. Section~\ref{sec:overview} provides an overview of the current state of research on LLMs in forecasting and anomaly detection. Section~\ref{sec:challenges} discusses the challenges and limitations associated with the application of LLMs in these domains. Section~\ref{sec:datasets} explores the datasets and data preprocessing techniques used in LLM-based forecasting and anomaly detection. Section~\ref{sec:evaluation-metrics} presents the evaluation metrics and methodologies used to assess the performance of LLMs in these tasks. Section~\ref{sec:forecasting} delves into the application of LLMs in forecasting, while Section~\ref{sec:anomaly-detection} focuses on their application in anomaly detection. Section~\ref{sec:threats} discusses the potential threats and risks associated with the use of LLMs in these domains. Section~\ref{sec:future} outlines the future directions and potential research avenues in the application of LLMs in forecasting and anomaly detection. Section~\ref{sec:related} provides an overview of related work, and Section~\ref{sec:conclusion} concludes the paper.
\looseness=-1

\section{Methodology}\label{sec:methodology}

In the rapidly evolving domain of artificial intelligence (AI), LLMs have emerged as copilot tools in various applications, notably in forecasting and anomaly detection~\cite{chen2022bert-log_anomaly,dang2020time_series}. However, despite their growing prominence, a substantial knowledge gap exists regarding their comprehensive capabilities and limitations in these contexts. This review is motivated by the necessity to consolidate and critically analyze the extant research concerning LLMs in these specific applications. In light of the rapid progress in model architectures and their diverse applications, this review aims to amalgamate knowledge on existing methodologies, performance evaluation metrics, and practical implementations while also identifying the prevailing challenges and limitations. This effort is imperative for both academic researchers and industry practitioners who aim to utilize these models effectively and serves as a foundational reference for future research and development in this field. By systematically examining and integrating diverse findings from recent studies, this review aims to offer a structured and comprehensive understanding of the current state-of-the-art, thereby guiding informed decision-making and strategic advancements in the application of LLMs.

In our study, we adopted the SLR methodology as outlined by Barbara Kitchenham \cite{kitchenham2012systematic_review, kitchenham2009systematic_literature}. This method is a comprehensive, rules-driven approach to finding and analyzing prior knowledge on a particular topic that involves a rigorous and transparent methodology to identify, evaluate, and interpret all available research relevant to a particular research question, topic area, or phenomenon of interest. It is designed to provide an exhaustive overview of the current state of research by integrating findings from various studies \cite{brereton2007lessons}. The SLR methodology is widely recognized and extensively applied in numerous academic surveys \cite{kitchenham2010systematic, aleti2012software, cocchia2014smart, chen2017science, liu2023pre-train_prompt}. The research questions (RQs) that guide our SLR process are presented below:

\begin{rqs}
  \item \textit{What methodologies are employed in LLMs for forecasting in different domains?} Different domains, such as finance, healthcare, weather, and retail, may require unique adaptations of LLMs to address domain-specific challenges and data characteristics. This question aims to explore and categorize the different methodologies and techniques used in LLMs for forecasting tasks, providing insights into their applicability across different sectors.
  \item \textit{How effective are LLMs in detecting anomalies compared to traditional anomaly detection methods?} Anomalies often exhibit distinct characteristics across diverse contexts, such as outlier financial transactions, atypical network traffic patterns, and unanticipated variations in health data. This question seeks to evaluate the performance and accuracy of LLMs in identifying outliers or unusual patterns in data, contrasting their effectiveness with that of conventional anomaly detection techniques.
  \item \textit{What are the limitations and challenges of using LLMs for forecasting and anomaly detection?} LLMs present a transformative potential for revolutionizing forecasting and anomaly detection due to their advanced pattern recognition and predictive capabilities. This question intends to identify the current limitations, challenges, and potential areas of improvement in using LLMs for these purposes, including factors like data prerequisites, computational expenditures, and model interpretability.
\end{rqs}

\textbf{RQ1} calls for a detailed exploration of the strategies, techniques, and
models used in applying LLMs across various sectors for predictive purposes.
\textbf{RQ2} seeks to evaluate and compare the performance of LLMs against
conventional techniques in identifying irregularities or unexpected patterns in
data.
\textbf{RQ3} necessitates a comprehensive exploration of the obstacles and constraints faced when employing these advanced models in specific predictive and analytical tasks.

After delineating the research questions, we strategically integrate various search engines and databases to identify pertinent studies, as outlined in Table~\ref{tab:search-source}. In order to find the most cutting-edge papers, we added OpenReview to the search for forthcoming papers that provide significant insight or data.

\begin{table}[!ht]
  \caption{Search Engines and Databases for Manual Search}
  \fontsize{8}{11}\selectfont
  \centering
  \begin{tabular}{ll}
    \toprule
    \textbf{Source} & \textbf{Search Scheme} \\
    \midrule
    Google Scholar                           \\(\href{https://scholar.google.com}{https://scholar.google.com}) & Full Text \\
    Web of Science                           \\(\href{https://www.webofscience.com}{https://www.webofscience.com}) & TS | TI | AB | AK | KP (Topic, Title, Abstract, Author Keywords, Keywords Plus) \\
    Scopus                                   \\(\href{https://www.scopus.com/}{https://www.scopus.com/}) & TITLE-ABS-KEY (Title, Abstract, Keywords) \\
    OpenReview                               \\(\href{https://openreview.net}{https://openreview.net}) & Keywords \\
    IEEE Xplore                              \\(\href{https://ieeexplore.ieee.org}{https://ieeexplore.ieee.org}) & Full Text \\
    ACM Digital Library                      \\(\href{https://dl.acm.org}{https://dl.acm.org}) & Title \\
    Springer Link                            \\(\href{https://link.springer.com}{https://link.springer.com}) & Full Text \\
    \bottomrule
  \end{tabular}
  \label{tab:search-source}
\end{table}

After retrieving studies through our established search strategy, we conducted a relevance assessment based on the inclusion and exclusion criteria outlined in Table~\ref{table:paper-criteria}. This process enabled the selection of primary studies offering direct evidence pertinent to the research questions.

\begin{table}[!ht]
  \caption{Paper Inclusion and Exclusion Criteria}
  \centering
  \begin{tabular}{l}
    \toprule
    \textbf{Inclusion criteria}                                                                        \\
    \midrule
    1) The paper claims the utilization of LLMs within the context of forecasting or anomaly detection \\
    2) The paper was published within a recent time frame of 3 years (i.e., \(\text{year} \ge 2020\))  \\
    3) The paper is peer-reviewed articles, conference proceedings, and academic journals              \\
    4) The paper was published in English with accessible full text                                    \\
    \midrule
    \textbf{Exclusion criteria}                                                                        \\
    \midrule
    1) Multiple publications reporting the same research or data                                       \\
    2) Published as a survey or literature review                                                      \\
    3) Short visionary paper, tool demo, and editorial                                                 \\
    4) Published in a workshop or a doctoral symposium                                                 \\
    5) Grey literature, non-peer-reviewed articles, or opinion pieces                                  \\
    \bottomrule
  \end{tabular}
  \label{table:paper-criteria}
\end{table}

\clearpage
\begin{table}[!ht]
  \caption{Overview of LLM-based Forecastor and Anomaly Detector Research}
  \fontsize{8}{11}\selectfont
  \centering
  \begin{tabular}{p{.05\linewidth}<{\centering\arraybackslash}p{.18\linewidth}p{.15\linewidth}p{.13\linewidth}p{.14\linewidth}p{.14\linewidth}}
    \toprule
    \textbf{Research}                        & \textbf{LLMs}                                   & \textbf{Task}                          & \textbf{Category}                                           & \textbf{Datasets}                                       & \textbf{Metrics}                      \\
    \midrule
    \cite{gruver2023large_language}          & GPT-3, GPT-4, Llama2-7b, Llama2-13b, Llama2-70b & Forecasting                            & Zero-shot                                                   & Darts, Monash, Informer                                 & MAE                                   \\
    \cite{zhou2023one_fits}                  & GPT-2, BERT                                     & Forecasting,\newline Anomaly Detection & Foundation Model                                            & ETT                                                     & MSE, MAE, MAPE, sMAPE                 \\
    \cite{shi2023language_models}            & GPT-3, GPT-3.5,\newline Llama2-13b              & Forecasting                            & Few-shot                                                    & ICEWS,\newline Amazon Review                            & RMSE                                  \\
    \cite{cao2024tempo_prompt-based}         & GPT-2                                           & Forecasting                            & Prompt-based                                                & ETT, Weather,\newline Electricity, TETS                 & MSE, MAE, sMAPE                       \\
    \cite{xue2023promptcast_a}               & BART, BigBird, Pegasus, GPT-3.5                 & Forecasting                            & Prompt-based                                                & CT, ECL, SG                                             & MAE, RMSE                             \\
    \cite{chen2022bert-log_anomaly}          & BERT                                            & Anomaly Detection                      & Foundation Model                                            & HDFS, BGL                                               & F1-Score                              \\
    \cite{lee2023lanobert_system}            & BERT                                            & Anomaly Detection                      & Foundation Model                                            & HDFS, BGL,\newline Thunderbird                          & F1-Score, AUROC                       \\
    \cite{dang2020time_series}               & BERT                                            & Anomaly Detection                      & Fine-tuning                                                 & KPI                                                     & F1-Score                              \\
    \cite{xue2022leveraging_language}        & BERT, RoBERTa, XLNet                            & Forecasting                            & Fine-tuning,\newline Foundation Model                       & SafeGraph                                               & RMSE, MAE                             \\
    \cite{ott2021robust_and}                 & BERT, GPT-2, XLNet                              & Anomaly Detection                      & Foundation Model                                            & OpenStack                                               & Precision, Recall, F1-Score           \\
    \cite{jin2024time-llm_time}              & Llama2-7b                                       & Forecasting                            & Prompt-based                                                & ETT, Weather,\newline Electricity, Traffic, ILI, M3, M4 & MSE, MAE, MAPE, sMAPE, MASE, OWA      \\
    \cite{li2022evaluating_bert}             & BERT                                            & Forecasting                            & Fine-tuning                                                 & SMD                                                     & MSE                                   \\
    \cite{zhang2023logprompt_a}              & BERT                                            & Anomaly Detection                      & Prompt-based                                                & HDFS, BGL                                               & Precision, Recall, F1-Score, Accuracy \\
    \cite{jin2021trafficbert_pre-trained}    & BERT                                            & Forecasting                            & Foundation Model                                            & METR-LA, PeMS-L, PeMS-Bay                               & RMSE, MAE, MASE, MAPE                 \\
    \cite{dang2021ts-bert_time}              & BERT                                            & Anomaly Detection                      & Fine-tuning                                                 & KPI, Yahoo                                              & Precision, Recall, F1-Score           \\
    \cite{huang2023improving_log-based}      & BERT                                            & Anomaly Detection                      & Fine-tuning,\newline Foundation Model,\newline Prompt-based & HDFS, BGL                                               & Precision, Recall, F1-Score           \\
    \cite{gupta2023learning_representations} & BERT                                            & Anomaly Detection                      & Few-shot,\newline Fine-tuning,\newline Zero-shot            & LFD, GSC, FC                                            & Precision, Recall, F1-Score           \\
    \cite{shao2022log_anomaly}               & BERT                                            & Anomaly Detection                      & Foundation Model                                            & HDFS, BGL,\newline Thunderbird                          & Accuracy, Recall, F1-Score            \\
    \cite{he2023parameter-efficient_log}     & BERT                                            & Anomaly Detection                      & Fine-tuning                                                 & BGL, Thunderbird, HDFS                                  & Precision, Recall, F1-Score           \\
    \cite{karlsen2023exploring_semantic}     & BERT                                            & Anomaly Detection                      & Foundation Model                                            & ECML/PKDD, CSIC, Apache                                 & F1-Score, Precision, Recall           \\
    \cite{dong2023simmtm_a}                  & Transformer                                     & Forecasting                            & Foundation Model                                            & ETT, Weather, Electricity,\newline Traffic              & MSE, MAE                              \\
    \cite{hu2023research_on}                 & BERT                                            & Anomaly Detection                      & Foundation Model                                            & HDFS, OpenStack                                         & Precision, Recall, F1-Score           \\
    \cite{almodovar2024logfit_log}           & BERT                                            & Anomaly Detection                      & Foundation Model                                            & HDFS, BGL,\newline Thunderbird                          & Precision, Recall, F1-Score           \\
    \cite{zhang2022logst_log}                & BERT                                            & Anomaly Detection                      & Foundation Model                                            & HDFS                                                    & Precision, Recall, F1-Score           \\
    \cite{le2021log-based_anomaly}           & BERT                                            & Anomaly Detection                      & Foundation Model                                            & HDFS, BGL,\newline Thunderbird, Spirit                  & Precision, Recall, F1-Score           \\
    \cite{huang2020hitanomaly_hierarchical}  & BERT                                            & Anomaly Detection                      & Foundation Model                                            & HDFS, BGL,\newline OpenStack                            & Precision, Recall, F1-Score           \\
    \bottomrule
  \end{tabular}
  \label{tab:papers-overview}
\end{table}

The selection process involved filtering papers published within a defined recent time frame to guarantee that the review accurately represented the current landscape of technology and research. This approach was adopted as studies outside the specified time frame may not accurately reflect current technologies and methodologies. Our selection criteria prioritized peer-reviewed articles, conference proceedings, and academic journals to maintain research credibility and rigor. In cases of multiple publications reporting identical research or data (e.g., a paper has an updated extended version), the most recent publication was chosen to eliminate redundancy.

Table~\ref{tab:papers-overview} provides a comprehensive overview of recent research studies focusing on the application of LLMs for forecasting and anomaly detection tasks. It systematically categorizes each piece of research according to the type of LLMs employed, the specific tasks addressed (forecasting, anomaly detection, or both), the methodological approach (e.g., zero-shot, few-shot, fine-tuning, foundation model, prompt-based), the datasets utilized in the studies, and the performance metrics used to evaluate the models' effectiveness.

The subsequent sections of this review delve into the detailed analysis of the methodologies, challenges, datasets, and performance metrics employed in LLM-based forecasting and anomaly detection. We also discuss the specific applications of LLMs in these domains, highlighting the current state of research, inherent challenges, and prospective future directions.

\section{Overview}\label{sec:overview}

The expansive domain of LLMs has ushered in unprecedented advancements in natural language processing, significantly impacting various tasks including forecasting and anomaly detection. This section provides a comprehensive overview of the current state and evolution of LLMs, delineating their foundational structures, development trajectories, and the pivotal role they play in transforming data analysis and predictive modeling. Beginning with a background on LLMs, we trace the evolution of language models from their nascent stages to the sophisticated pre-trained foundation models that serve as the backbone for contemporary applications. We then categorize tasks where LLMs have shown remarkable efficacy, specifically focusing on forecasting and anomaly detection, to illustrate the breadth of their applicability. Further exploration is dedicated to the diverse approaches employed to harness the power of LLMs, including prompt-based techniques, fine-tuning mechanisms, the utilization of zero-shot, one-shot, and few-shot learning, reprogramming strategies, and hybrid methods that combine multiple approaches for enhanced performance. This section aims to equip readers with a thorough understanding of the intricate landscape of LLMs, setting the stage for deeper exploration of their capabilities and applications in the subsequent sections.

\subsection{Background of Large Language Models}

In the evolution of language models, several iterative training paradigms have been applied. During the era of deep learning in NLP, models heavily relied on Long Short-Term Memory (LSTM)~\cite{yu2019review}, Convolutional Neural Network (CNN)~\cite{gu2018recent, ma2023implementation, zhu2019fully, yuan2023application}, and other deep models as feature extractors and \textit{Seq2Seq} was used as a basis for the framework, along with various modifications to the attention structures~\cite{shrestha2019review}. A key aspect of the technology was the design of intricate encoders and decoders. There was a marked gap between the effectiveness of NLP tasks and those in other domains, such as computer vision \cite{garderes2020conceptbert, quintana2022anterior}, and NLP research was in a lukewarm state, with a focus on intermediate task results such as tokenization \cite{ding2019towards}, part-of-speech tagging \cite{gui2017part}, and named entity recognition \cite{ritter2011named, li2020survey}.

The introduction of the Bidirectional Encoder Representations from Transformers (BERT)~\cite{devlin2019bert_pre-training} and Generative Pre-trained Transformer (GPT)~\cite{radford2018improving} has significantly propelled the advancement of the NLP field, leading to the widespread adoption of the Pre-training and Fine-tuning paradigm \cite{liu2023pre, gu2021domain, wu2019enriching, cui2021pre, bao2020will_go}. Large-scale corpora were utilized through task-oriented objectives, often referred to as unsupervised training (strictly speaking, it is supervised but lacks manually annotated labels) \cite{tao2018ruber, liao2022self}. Following this, fine-tuning on downstream tasks was implemented to enhance the final model's applicability. Notably, these models outperformed earlier deep learning methods~\cite{penedo2023refinedweb, li2023blip, kung2023performance, xiong2024decoding, ruan2024s2e}, prompting a focus on the meticulous design of pre-training and fine-tuning processes. The research tasks also shifted towards the ultimate goals of machine learning, such as text generation \cite{qu2023layoutllm, zhao2023more, li-etal-2023-synthetic, bao2020will_go}, dialogue system \cite{demetriadis2023conversational, hudevcek2023large, jing2024large}, machine translation \cite{sato2020vocabulary}, and others \cite{zhang2023optimizing}, with Pre-trained Languge Models autonomously learning the intermediate elements of the tasks.

For an extended period of time, Pre-trained Languge models based on BERT continued to receive most of the attention despite BERT and GPT series evolving in different directions \cite{clark2019does, jawahar2019does, zhang2020semantics, reimers2019sentence}. There were two primary reasons for this. To begin with, GPT creates greater difficulty when it attempts to predict the next context based on the preceding one, compared to BERT, which can detect both directions of context~\cite{radford2018improving}.
As a result, GPT series models were not as effective as BERT series models during the same period \cite{zhou2020evaluating}. Rather than aligning itself with a deliberate God's perspective, the GPT design pattern is more aligned with human learning strategies \cite{chan2023gpt}.
Secondly, GPT-3 represents the culmination of a process of gradual accumulation, and despite its impressive nature, its 175 billion parameters indicate a significant investment in training and usage~\cite{brown2020language}.
The high amount of investment in technology and funding did not result in a significant breakthrough in comparison to its potential benefits.
Consequently, it failed to capture the attention of AI researchers, let alone those in other industries.

The popularity of ChatGPT resulted in a surge of curiosity concerning the potential power of AI, marking a pivotal point in the progression of history.
The subsequent emergence of GPT-4, which demonstrated multimodal intelligence, prompted speculation as to whether the AGI era had arrived~\cite{openai2023gpt-4_technical}.
OpenAI has taken a unique approach in developing GPT, which principally concentrates on the `zero-shot' phenomenon from the Pre-trained Languge models era and implements prompt learning training methodologies that are more closely aligned with the trajectory of GPT~\cite{zhao2023explainability}. In order to achieve few-shot capabilities, the model size had to be increased, in-context learning had to be implemented and later, the future of artificial intelligence had to be considered. An integral aspect of this is the use of more human-friendly methods, which align with human ethics and common sense~\cite{dong2022survey}.
A major development within GPT was the introduction of supervised fine-tuning (SFT)~\cite{chen2020big} and the integration of human feedback (RLHF), aiming to align the model's knowledge with human knowledge~\cite{ouyang2022training}. This ultimately led to the development of ChatGPT\@.

In this section, we retrace and review the development trajectory of mainstream Large Language Models, from the first generation GPT-1 to GPT-4, marveling at the fact that the emergence of such powerful technologies does not occur overnight.

\subsubsection{Evolution of Language Models}

The journey of language models from simple rule-based systems to today's sophisticated LLMs represents a significant evolution in the field of NLP\@. This section delves into the chronological development of language models, highlighting key milestones and technological breakthroughs that have shaped their growth. Beginning with early statistical models that relied on N-gram probabilities, we trace the path towards the emergence of neural network-based models, which introduced a deeper understanding of context and semantics. The advent of transformer architectures marked a turning point, enabling models to process sequences of text with unprecedented efficiency and accuracy. We examine the transition from early transformers to the development of pre-trained foundation models, such as GPT-1 and BERT, which have set new standards for performance across a wide range of NLP tasks. This section not only charts the technological advancements that have propelled the evolution of language models but also sets the stage for understanding the current capabilities and limitations of LLMs in the broader context of forecasting and anomaly detection.

\bulletparagraph{Statistical language models}

Statistical Language Models (SLMs), developed in the 1990s, are based on statistical theories such as Markov Chains~\cite{mikolov2012statistical}.
These models use probabilistic methods to predict the next word in a sentence. The basic assumption behind SLMs is that the probability of each word depends only on the previous few words. This dependency length is fixed, forming the \(n\) in N-gram models. SLMs include Unigram, Bigram, and Trigram models, each with its unique operating principle \cite{pauls2011faster}:

\textbf{Unigram Model}: Each word in the text is independent of other words. Therefore, the likelihood of a sentence is calculated as the product of the probabilities of each word.

\textbf{Bigram Model}: The Bigram Model extends the concept of Unigram, assuming dependence on the previous word. Therefore, the likelihood of a sentence here is calculated as the product of the probabilities of each pair of consecutive words in the sentence.

\textbf{Trigram Model}: The Trigram Model takes this one step further, considering the probability of a word given its previous two words, thus creating a three-word context.

However, despite their simplicity and effectiveness, these models have limitations due to their design. Firstly, they encounter difficulties when dealing with contexts longer than the fixed length \(n\)~\cite{tremblay2011effects}. Secondly, they face challenges when dealing with high-dimensional data. As \(n\) increases, the number of transition probabilities grows exponentially, greatly reducing the accuracy of the model~\cite{wang2018sparse}.

To alleviate this problem, smoothing algorithms like Backoff Estimation and Good-Turing Estimation are used~\cite{sima2003backoff}.
When higher-order probabilities are unavailable, Backoff Estimation regresses to lower-order N-grams, effectively reducing the dimensionality. Conversely, Good-Turing Estimation adjusts the probability distribution for unseen events to deal with the problem of zero probability for unfamiliar word combinations - a problem known as data sparsity~\cite{orlitsky2015competitive}. While SLMs are computationally inexpensive, easy to implement, and interpretable, their inability to capture long-term dependencies and semantic relationships between words limits their use in complex language tasks~\cite{benjamin2012reconstructing}.

\bulletparagraph{Neural Network Language Model}

With the development of neural networks, Neural Network Language Models (NNLM) have demonstrated stronger learning capabilities than Statistical Language Models, overcoming the dimensionality disaster of N-gram language models and greatly improving the performance of traditional language models. The advanced structure of neural networks enables them to effectively model long-distance context dependencies.

The idea of training language models with neural networks was first proposed by Wei Xu and Alexander Rudnicky (2000)~\cite{xu2000can_artificial}. In their paper, they proposed a method of constructing a Bigram language model with neural networks. After that, the most classic work in training language models is proposed by Bengio~\textit{et al.} (2000)~\cite{bengio2000neural} published at NIPS\@. However, due to the difficulty of training neural network models, it wasn't until Bengio \textit{et al.} (2003)~\cite{bengio2003neural} proposed the Feed-forward Neural Network language model (FNNLM) that neural network language models aroused the interest of academia and industry. Subsequently, Mikolov~\textit{et al.} (2010)~\cite{mikolov2010recurrent} introduced recurrent neural networks (RNNs) into language modeling, greatly improving the performance of language models. Following this, improved versions of recurrent neural networks, such as Long Short Term Memory (LSTM) recurrent neural networks~\cite{hochreiter1997long} and Gated Recurrent Unit (GRU) neural networks~\cite{cho2014learning}, were successively used to further improve the performance of language modeling. In addition, convolutional neural networks~\cite{lecun1995convolutional, dauphin2017language} have unexpectedly achieved success in language modeling, and their performance can be compared with recurrent neural networks.

\textbf{FFNNLM}: Feed Forward Neural Network Language Models consist of three layers: the embedding layer, the fully connected layer, and the output layer~\cite{bengio2003neural}. The embedding layer maps the current word to a vector. It first obtains the \textit{n-1} words before the current word position, then obtains the word vectors of these \textit{n-1} words according to the word embedding matrix, and finally combines them together as the representation of the current word. It can be understood that the word embedding here is to get the representation of the \textit{n-1} words before the current word. The fully connected layer and the output layer receive the word vectors of the \textit{n-1} context words of the current word as input, and then predict the probability of the current word. By mapping words to a low-dimensional space for representation, the problem of sparsity can be solved, and the model has a certain generalization ability. However, this method still has certain defects. The first is the limitation of the context window, that is, the limitation of the \textit{n-1} context words related to the current word. In real scenarios, people's understanding of sentences does not have the restriction of only being able to see the previous \textit{n-1} words. Secondly, it does not take into account temporal information. The words in the sequence have a front-to-back relationship, but this method will ignore the temporal information and treat words at different positions uniformly~\cite{glorot2010understanding}.

\textbf{RNNLM}: Recurrent Neural Network Language Models were proposed as a solution to the window limitation issue~\cite{mikolov2010recurrent}. Using Recurrent Neural Networks, historical context information can be stored without being limited by the window length. The probability of the current word is calculated at each time step based on the current word and all previous contexts recorded by the RNN~\cite{sutskever2011generating, wang2023unveiling}. Even though RNN language models are capable of using unlimited context for prediction, the inherent challenges of RNNs make training the model quite challenging. It is common to encounter issues such as gradient vanishing or gradient explosion during the training process~\cite{hochreiter1998vanishing}. Consequently, a proposal was made to replace RNNs with Long Short-Term Memory networks.

\textbf{LSTM-RNN}: LSTM is a variant of RNN, and a more advanced RNN~\cite{hochreiter1997long}.
The essence of the algorithm remains the same, and it is capable of effectively processing sequence data~\cite{niu2023enhancing}. In RNN, the value of the hidden layer is stored at every moment and is used at the next moment to ensure that every moment contains information from the previous moment. We refer to the place where information about each moment is stored as a Memory Cell. The RNN stores all information as it does not have the capability of selecting information. The LSTM, however, is distinct because it is powerful and incorporates a gate mechanism. The LSTM consists of three additional gates that allow information to be selectively stored. As part of the information transmission process, the information is transmitted in the following order: the information is initially input through the input gate, then the forget gate determines whether the information has been forgotten in the Memory Cell, and finally it determines whether the information should be output at this time through the output gate.

Compared with the above three classic LMs, the performance of RNNLM (including LSTM-RNNLM) is superior to FFNNLM, and LSTM-RNNLM has always been the most advanced LM\@. The current NNLM is mainly based on RNN or LSTM\@. However, the representations of words learned by the previous models are unique and context-independent, which is evidently not the case in real-world situations. Language models should also allow words to learn related information from their contexts, as the same word can have different semantics in different contexts. Prior methods are unidirectional, that is, when calculating probability for the current word, only the information from the previous context is considered. In spite of this, people's understanding habits can be influenced by the context of the current semantics.

\textbf{ELMo:} Embeddings from Language Model refers to a deep contextualized word representation that simulates both the complexity of word forms and the variability of word form across linguistic contexts~\cite{peters-etal-2018-deep}. Multidirectional LSTMs are used in ELMo. This representation is comprised of the word vector of the word itself as well as the current state of the LSTM at the current word position. Bidirectional LSTM is used to capture context features, and the stacking of multiple layers of LSTM is used to enhance feature extraction capabilities. Because of the bidirectional nature, ELMo will divide the calculation of conditional probability into two parts, including using the previous context to calculate the probability of the current word and using the next context to calculate the probability of the current word. In the specific training process, the state of the last layer of LSTM is used to predict the probability of the word at the next position (whether forward or backward). In specific downstream tasks, the relevant word vectors obtained from the text through ELMo and the LSTM state values can be used as additional features of the current word to enhance the effect of downstream tasks through weighted averaging~\cite{sarzynska2021detecting}. This is also a typical feature-based pre-training model method.

\bulletparagraph{Attention Mechanism}

The attention mechanism was first proposed by Bahdanau \textit{et al.} (2014)~\cite{bahdanau2014neural}.The purpose of this mechanism is to address the bottleneck found in RNNs that only support fixed length inputs (as sentences grow longer, the amount of information that needs to be carried forward will also grow, so embeddings of fixed size may be insufficient). This paper proposes a structure for translation tasks in which the encoder in \textit{Seq2Seq} is replaced by a bidirectional recurrent network (BiRNN) and the decoding part is based on an attention model.

Since Attention Mechanism gives the model the ability to distinguish and identify, it is widely used in a variety of applications, including machine translation~\cite{luong2015effective}, speech recognition~\cite{chorowski2015attention}, recommender system~\cite{liu2024news, wang2020mrmrp, wang2021emrm} and image captioning~\cite{anderson2018bottom}. For example, in machine translation and speech recognition applications, different weights are assigned to each word in the sentence, making the learning of the neural network model more soft. At the same time, Attention Mechanism itself can serve as a kind of alignment relationship, explaining the alignment relationship between the input/output sentences in translation, and explaining what knowledge the model has learned.

The Attention Mechanism mimics the human visual and cognitive system~\cite{guo2022attention}, allowing neural networks to focus on relevant parts when processing input data. By introducing the attention mechanism, neural networks can automatically learn and selectively focus on important information in the input, improving the performance and generalization ability of the model. The attention mechanism is essentially similar to the human selective attention mechanism, and the core goal is also to select more critical information from a large amount of information. The most typical attention mechanisms include self-attention mechanism, spatial attention mechanism, and temporal attention mechanism. These attention mechanisms allow the model to allocate different weights to different positions in the input sequence, so as to focus on the most relevant part when processing each sequence element.

\textbf{Self-attention Mechanism:} Self-attention consists of the idea that when processing sequence data, each element is associated with other elements in the sequence, rather than solely dependent on its adjacent position~\cite{vaswani2017attention_is}. It adaptively captures the long-term dependencies between elements by calculating the relative importance between elements. Specifically, for each element in the sequence, the self-attention mechanism calculates its similarity with other elements and normalizes these similarities into attention weights. As a result of summing each element with its respective attention weight, the output of the self-attention mechanism can be determined.

\textbf{Multi-head Attention Mechanism:} Multi-head attention mechanism is developed based on the self-attention mechanism, which is a variant of the self-attention mechanism, aimed at enhancing the expressive power and generalization ability of the model~\cite{vaswani2017attention_is}. It uses multiple independent attention heads to calculate attention weights separately, and concatenates or sums their results to obtain richer representations.

\textbf{Channel Attention Mechanism:} This mechanism is based on calculating the importance degree of each channel; therefore, it is used frequently in convolutional neural networks~\cite{wang2020eca}. At present, the SENet model is considered the classic channel attention mechanism. SENet increases the network's ability to represent features by learning the relationship between channels (the importance of each channel), thus enhancing its performance.
Due to its spatial modeling capacity, this CBAM has been widely used in vision tasks~\cite{li2023bubble}.

\textbf{Spatial Attention Mechanism:} Spatial attention and channel attention both strive to accomplish the same goal in different ways~\cite{zhu2019empirical}. The channel attention algorithm is intended to capture the degree of importance of the channel whereas the spatial attention algorithm is intended to introduce an attention module that allows the model to learn the weights of the attention for different regions according to their importance. As a result, the model can pay more attention to important areas of the image and ignore areas of less importance. Among them, Convolutional Block Attention Module (CBAM), is the most typical. CBAM is a model designed to enhance the convolutional neural network's attention to images by combining channel and spatial attention~\cite{woo2018cbam}.

\bulletparagraph{Transformer}

Transformer was introduced in 2017, and its proposal attracted widespread attention to the Self-attention Mechanism, which further advanced the development of attention mechanisms~\cite{vaswani2017attention_is}.

In the past, the NLP field mainly relied on models such as recurrent neural networks (RNN) and convolutional neural networks (CNN) to process sequence data. However, these models often face problems such as gradient vanishing and low computational efficiency when dealing with long sequences~\cite{hochreiter1998vanishing}. The emergence of the Transformer broke this limitation. Transformer abandoned the traditional recursive structure and adopted a new self-attention mechanism to process sequence data in a more efficient and accurate way, enabling independent and parallel calculations at each position~\cite{vaswani2017attention_is}. The capabilities of this feature are aligned perfectly with those of modern AI accelerators, thus enhancing the efficiency of model computation. This innovation not only accelerates model training and inference but also opens up possibilities for distributed applications.

The self-attention mechanism is one of the core principles of the Transformer. It captures long-term dependencies by calculating the relationship between each element and other elements in the sequence. This mechanism allows the Transformer to compute in parallel when processing sequence data, greatly improving computational efficiency. At the same time, the self-attention mechanism can dynamically adjust weights according to different parts of the input sequence, making the model more flexible. Through the self-attention mechanism, the Transformer can perform parallel computations on each element in the input sequence and capture their relationships. This mechanism allows the model to better handle long sequence data and retain more context information during processing. In addition, the Transformer also uses techniques such as residual connections and normalization to effectively alleviate the problem of gradient vanishing and improve the training effect of the model.

As a revolutionary natural language processing model, the Transformer plays an important role in the field of artificial intelligence. It has pushed natural language processing to a new height and brought us great opportunities and challenges. The introduction of the Transformer model has changed the way traditional sequence models are processed and adopted the self-attention mechanism. Through the self-attention mechanism, the Transformer can capture long-term dependencies in the input sequence and better understand and generate natural language text. This revolution has enabled the Transformer to achieve outstanding performance in NLP tasks, with machine translation being the most outstanding representative. Translation models based on the Transformer, such as OpenAI's GPT~\cite{radford2018improving} and Google's BERT~\cite{devlin2019bert_pre-training}, have achieved unprecedented breakthroughs in translation quality, greatly improving the accuracy and fluency of translation. In addition, the Transformer has also shown strong capabilities in summarization and generation tasks~\cite{jin2023binary}, bringing us a more intelligent and natural interactive experience.

\subsubsection{Pre-trained Foundation Models}

Pre-trained foundation models have become the cornerstone of modern NLP, heralding a new era of language understanding and generation. This section explores the inception, development, and impact of these models, which are characterized by their vast knowledge bases, acquired through extensive pre-training on diverse and large-scale datasets. We delve into the mechanics behind their architecture, primarily focusing on transformer models such as GPT, BERT, and their successors, which have demonstrated remarkable versatility and performance across a multitude of NLP tasks. The discussion extends to the strategies employed in pre-training these models, including the objectives, datasets, and computational resources involved, as well as the challenges and ethical considerations arising from their deployment. Additionally, we explore how these foundation models serve as a platform for further fine-tuning and adaptation, enabling customization for specific tasks or domains, including forecasting and anomaly detection. By examining the pivotal role of pre-trained foundation models, this section aims to provide insights into their transformative potential in advancing the capabilities of large language models and their applications in real-world scenarios.

\bulletparagraph{BERT}

By introducing the bidirectional concept, Bidirectional Encoder Representations from Transformers (BERT) innovatively predicts both preceding and succeeding contexts~\cite{devlin2018bert}. As a pre-trained model, BERT significantly improves learning efficiency by requiring only a small number of parameters for fine-tuning in practical applications. In terms of structure, BERT is relatively simple, with Bert-Base and Bert-Large models composed of 12 and 24 repeated basic transformer blocks. The transformer block consists of three modules: Multi-Head Attention, Add\&Norm, and FFN\@. While the original transformer used triangular positional encoding~\cite{vaswani2017attention_is}, BERT adopts learnable positional encoding with a preset position count of 512, limiting the maximum sequence length to 512. BERT utilizes two unsupervised pre-training tasks: Masked LM, where some words are masked, and the network predicts their meaning based on context; and Next Sentence Prediction, a task determining whether two sentences are consecutive. It's worth noting that BERT encounters challenges in handling consecutive Mask Tokens and is not directly applicable to variable-length text generation tasks.

\bulletparagraph{GPT-1}

There has been a long history behind GPT-1 dating back to the groundbreaking paper \textit{``Attention is all you need''} \cite{vaswani2017attention_is}. According to it, Transformer is divided into two parts: encoder and decoder, both of which perform Multi-Head Self Attention, though the encoder is able to observe information from the entire source sequence while the decoder does not. The Bert model adapts the encoder, and when designing pre-training tasks, it predicts the missing intermediate words based on context, similar to filling in the blanks. Alternatively, GPT-1 utilises a decoder, which predicts the next context based on the previous context, thus allowing it to effectively perform masked multi-head self attention.

There are two stages in the PLM paradigm: \textit{pre-training} and \textit{fine-tuning}. The pre-training stage involves generating context predictions from a large-scale corpus of data. The fine-tuning stage involves training the model using downstream data and feeding the embedding of the last token into the prediction layer, which fits the label distribution of the downstream data. With an increase in layers, the accuracy and generalization capabilities of the model continue to improve, and further improvement is possible. Moreover, GPT-1 possesses an inherent capability for zero-shot learning, and this capability augments in tandem with the model's size. It is these two points that directly contribute to the emergence of subsequent GPT models.

\bulletparagraph{GPT-2}

GPT-2 is an enhanced version of GPT-1, based on the Transformer architecture for language modeling. GPT-2 can train models from massive unlabeled data, and the fine-tuning process enhances model performance, optimizing it for downstream tasks \cite{radford2019language_models}. In GPT-2, the language model is given greater emphasis in a zero-shot scenario, in which the model has not been trained or fine-tuned for downstream tasks prior to its application. A difference between GPT-2 and GPT-1 is that GPT-2 does not undergo fine-tuning for different tasks. Rather, it transforms the input sequences of downstream tasks. GPT-1 introduced special tokens, like start and separator symbols, but zero-shot scenarios prevent them from being used to fine-tune downstream tasks, as the model can't recognize these symbols without additional training. Therefore, in a zero-shot setting, input sequences for different tasks would be similar to the text seen during training, taking the form of natural language without task-specific identifiers.

The GPT-1 model is composed of 12 layers, whereas the BERT model is composed of 24 layers. However, the GPT-2 model consists of 48 layers with 1.5 billion parameters. The training data is derived from the WebText dataset, which undergoes some basic data cleaning. In accordance with the paper~\cite{radford2019language_models}, larger language models such as GPT-2 require more data to reach convergence, and experimental results indicate that current models are still underfitted. GPT-2 uses unidirectional transformers as opposed to BERT, which uses bidirectional transformers, and adopts a multitasking approach during the pre-training stage. Rather than learning on a single task, it learns across multiple tasks, which ensures that the losses of each task converge. The main transformer parameters are shared across different tasks. MT-DNN \cite{liu2019multi-task_deep} was the inspiration for this approach, which further enhanced the generalization ability of the GPT-2 model. As a result, GPT-2 exhibits impressive performance even without fine-tuning.

GPT-2 performs better than unsupervised algorithms in many tasks, showcasing its zero-shot capabilities. However, it still exhibits some deficiencies compared to fine-tuning algorithms with supervised feedback.

\bulletparagraph{GPT-3}

GPT-3 maintains the concept of excluding fine-tuning and focusing solely on a universal language model, as does the previous model. However, there are some technical replacements: GPT-3 introduces the sparse attention module from Sparse Transformer, aimed at reducing computational load~\cite{brown2020language_models}. It is necessary to make this adaptation since GPT-3 has increased the parameter size even further compared to GPT-2, reaching a staggering number of 175 billion parameters. When it comes to downstream tasks, GPT-3 utilizes a few-shot approach without fine-tuning, highlighting substantial differences in accuracy between varying parameter magnitudes and showing the extraordinary capabilities of large models.The training data for GPT-3 includes the Common Crawl dataset for lower quality, as well as the WebText2 dataset for higher quality, as well as the Books1, Books2, and Wikipedia datasets for higher quality~\cite{brown2020language_models}. GPT-3 assigns different weights to datasets according to their quality, with higher-weighted datasets being more likely to be sampled during training.

Additionally, according to the paper~\cite{brown2020language_models}, the one-shot effectiveness shows a significant and noticeable improvement when applied to large language models. As prompts are added, this improvement is further amplified, while marginal returns for few-shot decline gradually as prompts are added. Prompts are evident up to a point around eight-shot, but once eight-Shot is reached, their influence diminishes. Prompts are effectively ineffective beyond ten-shots. GPT-3 differs from previous models in that it is capable of achieving few-shot capabilities through constructing prompts. This capability is referred to as In Context Learning. Even though both fine-tuning and In Context Learning may appear to provide examples of large language models, they are fundamentally different. During fine-tuning, downstream tasks are performed, examples are provided, and parameter gradients are updated. As opposed to this, In Context Learning focuses on downstream tasks using examples without updating the parameters.

GPT-3 has the advantage of a high degree of generalization. The model can perform various subtasks without having to be fine-tuned because natural language instructions can be included in the input sequence without requiring any adjustments. GPT-3 achieves or exceeds state-of-the-art in some tasks, thus confirming that larger model sizes are associated with higher task performance. The few-shot capability of GPT-3 is more powerful than both one-shot and zero-shot capabilities in most situations. Moreover, the authors anticipate that GPT-3 could potentially have societal implications~\cite{brown2020language_models}. For instance, it has the potential to facilitate the generation of fake news, spam, and academic papers. Given the likely influence of the biases present in GPT-3's training data, namely racial, religious, and gender biases, the generated text might mirror these issues.

\bulletparagraph{InstructGPT (GPT-3.5)}

According to the paper~\cite{ouyang2022training_language}, the InstructGPT model is proposed to enhance the alignment between the outputs of the model and the user's intentions. Despite GPT-3's remarkable capabilities in diverse NLP tasks and text generation, it can still generate inaccurate, misleading, and harmful information that can negatively impact society. Moreover, GPT-3 often does not communicate in a form that is accepted by the human audience. Consequently, OpenAI introduces the concept of ``Alignment'', which strives to align model outputs with human preferences and intentions.

InstructGPT defines three key objectives for an idealized model of language: helpful, honest, and harmless~\cite{ouyang2022training_language}. InstructGPT requires two rounds of fine-tuning for its model: from GPT-3 to SFT (supervised fine-tuning), and then to RL (reinforcement learning). As a result of the SFT model, it can be addressed the problem of GPT-3's inability to guarantee answers based on human instructions, to be helpful, and to generate safe responses without the need for manual annotation data to refine the answers. Using the reward model, a ranking-based discriminative annotation is introduced, which is much less costly than generative annotation. Furthermore, through the use of reinforcement learning capabilities, the model is able to gain a deeper understanding of human intentions.

When comparing InstructGPT to GPT-3, several advancements can be observed. It has the capacity to comprehend user instructions, explicit or implicit, encompassing goals, constraints, and preferences, subsequently generating outputs that align more closely with user expectations and needs. InstructGPT is capable of more effectively utilizing information or structures provided in prompts, and can make reasonable inferences or creations based on that information. By consistently maintaining output quality, errors or failures can be reduced. Furthermore, InstructGPT, with 13 billion parameters, significantly outperforms GPT-3, which consists of 175 billion parameters.

\bulletparagraph{GPT-4}

According to the paper and experiments~\cite{openai2023gpt-4_technical}, GPT-4 significantly improves the GPT model scale and training methods, while having over a trillion parameters compared to GPT-3. By utilizing a novel training technique, known as Reinforcement Learning from Human Feedback (RLHF), the GPT-4 model is able to generate text in a more natural and accurate manner. RLHF utilizes a combination of pre-training and fine-tuning strategies, engaging in interactive conversations with human operators in order to train through reinforcement learning. This enhances GPT-4's understanding of context and questions and improves its performance on specific tasks~\cite{katz2023gpt, nori2023capabilities, wang2023knowledge}. In general, GPT-4 follows the same training strategy as ChatGPT, based on the principles of pre-training, prompting, and prediction.

GPT-4 introduces three significant enhancements: 1. The implementation of a rule-based reward model (RBRM); 2. Integration of multi-modal prompt Learning to support various prompts; 3. Incorporation of a chain of thought mechanism to enhance overall coherence in thinking. According to the paper~\cite{openai2023gpt-4_technical}, GPT-4 is a robust multimodal model able to process both image and text inputs, generating text outputs that are ranked in the top 10\% of test takers. Compared to GPT-3.5, which falls within the bottom 10\%, this is a significant improvement. The GPT-4 language model outperforms many state-of-the-art NLP systems on traditional benchmarks~\cite{liu2023gpt, chang2023survey, rathje2023gpt}. Specifically, the report addresses a key project challenge involving the development of deep learning infrastructure and optimization methods that exhibit predictable behavior across a wide range of scales. Additionally, it discusses interventions implemented to address potential risks associated with GPT-4 deployment, such as adversarial testing with domain experts and the implementation of a model-assisted safety pipeline.

Since ChatGPT-3 and GPT-4 are trained on large amounts of text from the internet, they may be subject to biases and inaccurate information. The OpenAI team has implemented additional filters in GPT-4 to address this issue, reducing the likelihood of inappropriate content being generated and improving control over the generated text~\cite{openai2023gpt-4_technical}. The GPT-4 has a number of challenges and issues, however, it demonstrates considerable potential in several different application scenarios, opening up a wide range of possibilities for the development of artificial intelligence.

\bulletparagraph{AI21 Jurassic-2}

According to the document in the website~\cite{Jurassic2}, Jurassic-2, a customizable language model designed to power natural language use cases, is considered one of the largest and most complex models in the world. Jurassic-2, developed based on Jurassic-1, includes three base models in different size: Large, Grande, and Jumbo. In addition to comprehensive enhancements in text generation, API latency, and language support, Jurassic-2 also opens up command fine-tuning and data fine-tuning to help businesses and individual developers create customized ChatGPT assistants.

Certain types of specific fine-tuning are realized in Jurassic-2. To perform semantic search, Jurassic-2 understands the intent and context of queries and retrieves relevant text snippets from documents. As part of its context-based Q\&A service, Jurassic-2 provides answers based solely on specific contexts, with automatic retrieval from document libraries. When it comes to summarizing content, it can be used to obtain documents (original texts or URLs) and provide key points within them. According to the input requirements of the user, the obtained text can be output in a specific style, etc., resulting in nine fine-tuning options.

\bulletparagraph{Claude}

According to the website introduction~\cite{claude}, Claude is an artificial intelligence assistant developed by Anthropic with a cheerful personality and a rich individuality, designed to provide users with accurate information and answers. Anthropic was established in 2021, co-founded by several former OpenAI members, including Dario Amodei, Daniela Amodei, Tom Brown, Chris Olah, Sam McCandlish, Jack Clarke, and Jared Kaplan. They have rich experience in the field of language models and have participated in the development of models such as GPT-3. Google is the main investor of the company, having invested 300 million dollars in it.

There is not much information available as of yet, but Anthropic's research paper mentions AnthropicLM v4-s3 as a 52-billion-parameter model that has already been trained~\cite{bai2022constitutional}. The model is an autoregressive one trained on a large text corpus unsupervised, similar to the GPT-3 model. To generate fine-tuned outputs, Anthropic uses a unique process known as \textit{``Constitutional AI''}, which uses a model rather than humans. Anthropic names it \textit{``Constitutional AI''} because they began with a list of ten principles that constituted a ``constitution''. Despite not being publicly disclosed, Anthropic says its principles are based on beneficence (maximizing positive impact), nonmaleficence (avoiding giving harmful advice) and autonomy (respecting freedom of choice).

\bulletparagraph{BLOOM}

BLOOM, an acronym for BigScience Large Open-science Open-access Multilingual Language Model, is a language model possessing 176 billion parameters that has been trained on 59 natural languages and 13 programming languages. The model was trained on \textit{Jean Zay}, a supercomputer funded by the French government and managed by \textit{GENCI} (donation number \textit{2021-A0101012475}), installed at the National Computing Center IDRIS of the French National Center for Scientific Research (CNRS)~\cite{workshop2023bloom}.

Each component of BLOOM was carefully designed, including the training data, the model architecture and the training objectives, as well as the engineering strategies for distributed learning. BLOOM was trained based on modifications to Megatron-LM GPT2, using Megatron-DeepSpeed for training. This model is divided into two parts: Megatron-LM provides Transformer implementation, tensor parallelism, and data loading primitives, while DeepSpeed provides ZeRO optimizer, model pipeline, and general distributed training components~\cite{workshop2023bloom}. As a rule of thumb, it mainly utilizes the decoder-only structure, normalization of the word embedding layer, linear bias attention position encoding with GeLU activation function, etc. Currently, it is the largest open-source language model in the world, and it is transparent in many ways, disclosing the materials used for training, the difficulties encountered during development, and the methods for evaluating its performance.

In addition, it is also important to note that the BLOOM model is subject to the same disadvantages as other large language models, in the sense that inaccurate or biased language may be hidden. On the one hand, the project adopts the new \textit{``Responsible AI License''} in order to avoid being applied to high-risk areas such as law enforcement or healthcare, and it is also prohibited from use in harm, deception, exploit, or impersonation. On the other hand, Hugging Face believes that open source will enable the AI community to contribute to the improvement of this model.

\bulletparagraph{Hugging Face}

Hugging Face is a platform that focuses on natural language processing (NLP) and artificial intelligence (AI). The platform currently hosts over 320,000 models and 50,000 datasets, allowing machine learning practitioners around the world to collaborate on developing models, datasets, and applications~\cite{wolf2020huggingfaces}. Its abundant repository of pre-trained models and codes is widely used in academic research. It helps people keep track of popular new models and provides a unified coding style to use various different models such as Bert, XLNet, and GPT\@. Its Transformers library has also been open-sourced on GitHub~\cite{wolf-etal-2020-transformers}, which provides pre-trained models and fine-tuned models for different tasks. The Hugging Face website allows users to compare models easily, and they can find a pre-trained model and train it using their own data. Whatever the task is, Hugging Face provides the appropriate models and tools. This includes text classification, question answering systems, machine translation, text generation, or sentiment analysis. Developers are thus able to quickly and effectively build and deploy NLP solutions across a variety of applications.

Hugging Face not only provides a wide range of pre-trained models, but it also supports customization and extension. Developers can adjust the model according to their specific needs, or further train on the basis of existing models.

\subsection{Task Categorization}

The versatility of LLMs is showcased through their application across a diverse range of tasks, each presenting unique challenges and opportunities for innovation. This section categorizes and examines the specific roles LLMs play in two critical areas: forecasting and anomaly detection. In forecasting, we explore how LLMs contribute to predicting future events, trends, and behaviors, leveraging historical data and linguistic patterns to generate insights with significant accuracy. Anomaly detection, on the other hand, highlights the models' ability to identify outliers or unusual patterns within data, which is pivotal for security, quality control, and operational efficiency. Through a detailed exploration of these tasks, we aim to elucidate the methodologies and approaches employed by LLMs, ranging from direct application in a zero-shot or few-shot context to more complex fine-tuning and hybrid strategies. This section not only underscores the broad applicability of LLMs but also sets the stage for a deeper dive into the specific techniques and challenges associated with each task, providing a structured framework for understanding the multifaceted impact of language models in contemporary computational linguistics and data analysis domains.

\subsubsection{Forecasting}

The work of large language models in time series forecasting can basically be divided into two types. One method involves applying large language models directly to time series predictions, focusing on converting time series data into input data suitable for the models, such as GPT, Llama, and others. Another type is to train a large language model in the domain of time series, by using a large amount of data from several time series datasets to jointly train a large language model in the domain of time series, which can then be used for downstream time series tasks.

Specifically, the paper focuses on the second type, examining the ways in which researchers train large language models across a variety of domains.

\subsubsection{Anomaly Detection}

Anomaly detection can be divided into two categories.  In the first category of anomaly detection, training data with labels are provided, and a classifier is first trained using these data, and there is no ``unknown'' in the data and labels. However, it is expected that the classifier will be able to determine that the newly acquired training data differ from the original training data and label the new training data as ``unknown''. This is also known as Open-set Recognition. A second category consists of all training data that are unlabeled and anomalies are determined based on similarity between the data. The second category includes two situations: clean data, which means that all data is normal data, and polluted data, which means that some abnormal data has been mixed in with the training data.

Specifically, the paper focuses on the second subcategory of the second type of anomaly detection, examining the ways in which researchers train large language models across a variety of domains.

\subsection{Approaches}

The application of LLMs across various tasks, including forecasting and anomaly detection, involves a spectrum of innovative approaches, each tailored to optimize performance and accuracy. This section delves into the core methodologies employed to leverage LLMs, presenting a comprehensive overview of the strategies that have emerged as most effective in harnessing their potential. We begin with prompt-based methods, which involve crafting input prompts that guide the model toward generating desired outputs, demonstrating the flexibility and creativity inherent in interacting with LLMs. The discussion then moves to fine-tuning, a process of adjusting a pre-trained model's parameters to better suit specific tasks or datasets, enhancing its applicability and precision. The exploration of zero-shot, one-shot, and few-shot learning highlights how LLMs can perform tasks with minimal to no task-specific data, showcasing their remarkable adaptability. Reprogramming introduces the concept of modifying input data in ways that exploit the model's latent knowledge without altering its parameters, offering an innovative angle on model utilization. Lastly, hybrid approaches that combine multiple techniques are examined, illustrating the dynamic and evolving landscape of LLM application methods. This section aims to provide a thorough understanding of the diverse approaches to deploying LLMs, paving the way for their effective use in addressing complex challenges in NLP and beyond.

\subsubsection{Prompt-based}

Prompt-based refers to the transformation of input text information according to a specific template, restructuring the task into a form that can make full use of pre-trained language models~\cite{shin2020autoprompt}.

Different from traditional supervised learning, Prompt-based learning directly utilizes language models pre-trained on a large amount of raw text, and by defining a new prompt function, allows the model to perform few-shot or even zero-shot learning, adapting to new scenarios with only a small amount of annotated data or no annotated data.Unlike traditional fine-tuning methods, prompt learning adapts to various downstream tasks based on language model methods, usually without the need for parameter updates.

\subsubsection{Fine-tuning}

Fine-tuning fundamentally involves the transformation of general-purpose models into specialized ones. It entails taking pre-trained models and further training them on smaller, specific datasets to refine their capabilities and enhance their performance in a particular task or domain. This process serves as a bridge between generic pre-trained models and the unique requirements of specific applications, ensuring that the language model aligns closely with human expectations.

The procedure of fine-tuning is more resource-efficient and cost-effective in comparison to training a model from scratch. The latter necessitates extensive text datasets, significant computational resources, and substantial financial investment. In contrast, fine-tuning involves the adaptation of a pre-trained model to a smaller, task-specific dataset, which necessitates fewer resources, less time, and less financial investment.

\subsubsection{Zero-shot, One-shot, and Few-shot}

Zero-shot is a machine learning paradigm where the model is capable of making predictions about unseen classes without explicit training on these classes, this approach is widely used in industry research~\cite{song2023zeroprompt}. This is achieved by leveraging the model’s understanding of other, analogous classes to infer characteristics of the new classes. For instance, consider a model trained on a dataset of various types of birds. This model could be utilized to predict new bird species, such as sparrows and eagles, without explicit training on these species. This is possible because the model understands that all birds share certain common characteristics, such as feathers, beaks, and wings, which allows it to make educated guesses about the new bird species.

The one-shot is a machine learning paradigm where the model is capable of making predictions about new classes after being trained on a single instance of that class~\cite{song2023zeroprompt}. This task is more challenging than zero-shot learning, as the model has limited data to work with.
For example, a model trained on a dataset of various types of flowers could be used to predict a new flower, such as a daisy, after being trained on a single image of a daisy. The model can use this image to learn about the daisy's features, such as its petals, stem, and leaves.

Few-shot is a machine learning paradigm that lies between zero-shot and one-shot. In few-shot, the model is trained on a handful of examples from each new class. This task is more challenging than one-shot but less so than zero-shot~\cite{gao2020making}. For instance, a model trained on a dataset of various types of trees could be used to predict a new type of tree, such as a Japanese maple tree, after being trained on a few images of Sugar maple trees, Norway maple trees, and Field maple trees. The model can use these images to learn about the maple tree's features and make inferences about how the maple tree is similar to and different from other types of trees.

\subsubsection{Reprogramming}

Model Reprogramming, alternatively referred to as Adversarial Reprogramming \cite{elsayed2018adversarial}, represents a burgeoning field within Machine Learning. This approach involves repurposing an existing model for a novel task, circumventing the necessity for retraining, or fine-tuning the original model. Instead, the methodology modifies the inputs of the model to facilitate its application to a new adversarial task. Given that Model Reprogramming incurs a lower computational cost and necessitates less access to the model parameters in comparison to retraining or fine-tuning, it has been successfully extended for applications such as domain adaptation \cite{chen2022model}, knowledge transfer, and bias elimination in models \cite{zhang2022fairness}.

\subsubsection{Hybrid}

Hybrid methodologies amalgamate the strengths of diverse approaches to augment the performance and versatility of LLM models. Typically, these methodologies incorporate both rule-based and machine learning methods, capitalizing on the benefits of each. The rule-based approaches are reliant on pre-established linguistic rules and knowledge graphs, offering an explicit representation of knowledge with rich, expressive, and actionable descriptions of concepts. The machine learning approaches employ statistical techniques to learn from data. They are particularly adept at managing large-scale, complex tasks where manually crafting rules would be impractical. Hybrid approaches have also been extended for a variety of applications. They present a promising direction for enhancing the capabilities of LLMs, empowering them to handle more complex tasks and adapt to new domains effectively.

\section{Challenges}\label{sec:challenges}

In the realm of forecasting and anomaly detection, the deployment of LLMs represents a paradigm shift towards leveraging vast amounts of data for predictive insights. However, this approach is fraught with significant challenges that stem from the inherent properties of time series data, the lack of labeled instances, the prevalence of missing values, and the complexity of processing noisy and unstructured text data. These obstacles necessitate a sophisticated understanding and innovative methodologies to harness the full potential of LLMs in these applications.

The intricate nature of time series data, characterized by complex seasonality and patterns, demands models capable of capturing and forecasting dynamic temporal behaviors. This complexity is compounded by the multifaceted influences affecting time series, including but not limited to economic indicators, weather conditions, and social events, which introduce additional layers of difficulty in modeling efforts. Moreover, the scarcity of labeled data, especially in the context of anomaly detection, poses a significant hurdle. The effectiveness of LLMs in such scenarios is contingent upon developing and applying advanced strategies that can leverage limited annotations to discern patterns indicative of anomalies. Another pervasive issue in time series analysis is the occurrence of missing data, a consequence of various disruptions in data collection and transmission processes.
Different from the computer vision models that can be trained from a small amount of data~\cite{bateni2020improved,Wang2023AcceleratingTC}, LLMs require huge natural language corpora for training.
Addressing this challenge requires robust imputation methods that can seamlessly integrate with LLMs to ensure the integrity and continuity of the data being analyzed. In order to get reproducible and reusable datasets for analytics, the cORe~\cite{core} platform can be exploited.
Furthermore, the analysis of unstructured text data introduces additional complexity, as such data often contain high noise and irrelevant information. Effective preprocessing and feature extraction methods are imperative to distill valuable insights from unstructured text, necessitating a nuanced approach to understanding and extracting pertinent information.

These challenges underscore the necessity for innovative solutions that adapt to the complexities of time series data and unstructured text, ensuring that LLMs can be effectively applied to forecasting and anomaly detection tasks. The development of such solutions remains an active area of research, with the potential to significantly advance predictive analytics capabilities.

\subsection{Complex Seasonality and Patterns}

The challenge of modeling complex seasonality and patterns in time series data is a formidable obstacle in the application of LLMs to forecasting and anomaly detection tasks. Time series data can exhibit a wide range of seasonal behaviors, from simple annual cycles to intricate patterns that span multiple temporal resolutions, such as daily, weekly, and monthly fluctuations. These patterns may also interact with each other, creating complex seasonal dynamics that are difficult to predict.

One of the primary challenges in addressing complex seasonality is the requirement for LLMs to not only recognize these patterns but also to understand their underlying causes and interactions. Traditional models might struggle to capture such complexities without significant customization or the inclusion of domain-specific knowledge. With their vast parameter spaces and deep learning capabilities, LLMs offer a potential solution to this problem by learning from large datasets encompassing the full range of seasonal variations and their associated factors. However, this requires a substantial volume of high-quality, granular data spanning multiple seasonal cycles to train these models effectively.

Moreover, the presence of external factors such as holidays, economic fluctuations, and weather conditions further complicates the modeling of seasonality. These factors can introduce additional variance into the time series, making it challenging to isolate and predict the impact of seasonality on the data. For LLMs to accurately forecast under these conditions, they must be capable of integrating external data sources and contextual information into their predictions. This requires advanced data processing capabilities and the ability to infer causal relationships and adapt to changing conditions over time.

Another aspect of complexity arises from the non-linear interactions between different seasonal patterns. For instance, the effect of a holiday on consumer behavior might vary significantly depending on the day of the week it occurs or its proximity to other events. Capturing such non-linearity and interactions is crucial for accurate forecasting and anomaly detection, demanding sophisticated modeling techniques that can account for a wide range of dependencies and conditional effects.

Addressing complex seasonality in time series data with LLMs requires not only extensive training data but also advanced optimization techniques. Stochastic optimization methods \cite{decision, emsemble, compromise}, including multi-stage stochastic programming and stochastic integer programming, play a pivotal role in enhancing LLMs' ability to capture intricate patterns and variations inherent in temporal dynamics. These approaches introduce flexibility and adaptability, allowing the model to make sequential decisions over different time horizons and incorporate discrete variables, thereby improving its performance in forecasting and anomaly detection tasks amidst complex seasonal behaviors. The synergy between deep learning capabilities and stochastic optimization equips LLMs to recognize, understand, and adapt to diverse temporal patterns, emphasizing the importance of careful parameter tuning for optimal performance across various time series scenarios.

In summary, addressing the challenge of complex seasonality and patterns in time series data with LLMs involves a multifaceted approach that includes the development of models capable of learning from large and diverse datasets, the integration of external factors and contextual information, and the ability to model non-linear interactions and dependencies. Success in these endeavors can significantly enhance the accuracy and reliability of forecasting and anomaly detection, unlocking new possibilities for predictive analytics in various domains.

\subsection{Label Deficiency}

The issue of label deficiency represents a significant challenge in the deployment of LLMs for forecasting and anomaly detection tasks, particularly in domains where labeled data are scarce or expensive to obtain. This scarcity is acutely felt in anomaly detection, where anomalous events are inherently rare and thus less likely to be represented in training datasets. The lack of labeled examples hampers the ability of models to learn the nuanced patterns that differentiate normal from anomalous behavior, leading to decreased accuracy and increased false positives or negatives.

In the context of forecasting, the challenge of label deficiency arises from the need to train models on historical data that may not contain explicit labels for future events or outcomes. While some forecasting tasks may have access to labeled data for past time periods, the absence of labels for future time points makes it difficult to evaluate the accuracy of predictions and to train models on the specific patterns associated with future events.

Several strategies have been proposed and adopted within the machine learning community to combat label deficiency. One such strategy involves using semi-supervised learning techniques, which allow models to learn from labeled and unlabeled data. This approach leverages the abundant unlabeled data to improve model generalization, thereby mitigating the effects of limited labeled data. With their capacity to understand and generate human-like text, LLMs can be particularly adept at exploiting the context provided by unlabeled data to infer underlying patterns and relationships.

Data augmentation is another critical strategy for addressing label deficiency. By artificially augmenting the dataset with synthetic examples through techniques like oversampling, undersampling, or generating new instances via transformations, models can be exposed to a broader range of scenarios than those represented in the original labeled dataset. This exposure helps improve the robustness and generalizability of the model. However, generating realistic and relevant synthetic data that accurately captures the complexity of real-world scenarios is challenging and requires sophisticated approaches.

Transfer learning has also emerged as a potent solution to the challenge of label deficiency. By pre-training models on large, diverse datasets and then fine-tuning them on the target task with limited labeled data, LLMs can leverage learned representations and knowledge to enhance their performance on tasks with scarce labels. This approach is particularly effective in domains where pre-trained models have been exposed to relevant contexts or languages during their initial training phase.

Despite these strategies, the challenge of label deficiency remains a significant barrier to the effective application of LLMs in forecasting and anomaly detection tasks. The development of more advanced techniques for semi-supervised learning, data augmentation, and transfer learning continues to be a crucial area of research. Additionally, exploring innovative ways to leverage unlabeled data, such as unsupervised anomaly detection methods that do not rely on labeled examples, may offer new pathways to overcoming the limitations imposed by label scarcity.

\subsection{Missing Data in Time Series}

Addressing missing data in time series is a critical challenge when applying LLMs for forecasting and anomaly detection. Missing data can arise from many sources, including equipment malfunctions, data transmission errors, or simply gaps in data collection. These missing values pose a significant problem, as they can lead to inaccuracies in predictions and analyses if not properly handled. The issue is further complicated by the sequential nature of time series data, where the temporal dependencies and patterns play a crucial role in forecasting and anomaly detection tasks.

One common approach to managing missing data is through imputation, where missing values are filled in based on available data. The complexity of imputation varies with the amount and type of data missing, as well as the patterns and dependencies present in the time series. Simple imputation methods, such as mean or median imputation, are often inadequate for time series data due to their inability to capture temporal dynamics. More sophisticated techniques, such as linear interpolation or time series-specific methods like ARIMA-based imputation, can provide better results by leveraging the temporal structure of the data. However, these methods may still fall short when dealing with non-linear patterns or long gaps of missing data.

LLMs offer promising avenues for addressing the challenges of missing data through their ability to model complex patterns and relationships in data. By training on large datasets, LLMs can learn the underlying structures and dependencies in time series, potentially enabling them to predict missing values with higher accuracy than traditional methods. Moreover, LLMs can incorporate contextual information and external variables, providing a more nuanced approach to imputation that considers both temporal dynamics and external influences.

Despite the potential of LLMs to handle missing data, several challenges remain. Ensuring the quality and reliability of imputed values is paramount, as inaccuracies can propagate through subsequent analyses and lead to misleading conclusions. Furthermore, the computational complexity of using LLMs for imputation can be significant, particularly for large datasets with extensive missingness. There is also the need for careful model tuning and validation to avoid overfitting and ensure that the imputation method generalizes well across different time series.

In summary, while LLMs present a promising solution to the challenge of missing data in time series, their effective application requires careful consideration of the methods used for imputation, the potential for model overfitting, and the computational demands of the task. Ongoing research into more advanced imputation techniques and the development of LLMs designed explicitly for time series data will be crucial in overcoming these challenges and unlocking the full potential of LLMs in forecasting and anomaly detection.

\subsection{Noisy and Unstructured Text Data}

The challenge of noisy and unstructured text data is particularly pronounced in applications involving LLMs for forecasting and anomaly detection. Unstructured text, which includes various formats such as social media posts, news articles, and log files, often contains a significant amount of noise—irrelevant information, typos, slang, and ambiguous expressions that can obfuscate meaningful insights. This noise complicates the task of extracting valuable features and patterns that are critical for accurate predictions and anomaly identification.

To effectively harness the power of LLMs in processing noisy and unstructured text data, a comprehensive approach to data preprocessing is essential. This involves cleaning the data by removing or correcting typos, standardizing terminology, and filtering out irrelevant information. Such preprocessing steps are crucial for reducing the noise in the data and making it more amenable to analysis by LLMs. However, the challenge lies in executing these steps without losing important contextual or nuanced information that may be crucial for the task at hand.

Beyond preprocessing, feature extraction from unstructured text represents another significant challenge. Traditional methods may not fully capture the complexity and richness of the data, limiting the model's ability to understand and predict based on the text. LLMs, with their advanced natural language processing capabilities, offer a promising solution by automatically identifying and extracting relevant features directly from text. They can discern patterns, sentiments, and relationships that are not immediately apparent, providing a deeper understanding of the data. However, leveraging LLMs for feature extraction from noisy and unstructured text also requires careful model tuning and validation. The models must be trained on sufficiently diverse datasets to ensure they can generalize well across different types of text and noise levels. Moreover, there is a need for mechanisms to assess the relevance and importance of the extracted features, as not all information gleaned from the text may be useful for forecasting or anomaly detection purposes.

Incorporating external knowledge bases and ontologies is another strategy that can enhance the performance of LLMs in dealing with unstructured text. By providing additional context and background information, these resources can help the model disambiguate and interpret complex or ambiguous text more effectively. However, integrating such external sources into the modeling process introduces additional complexity and raises questions about the scalability and adaptability of the solution.

In conclusion, while noisy and unstructured text data presents a significant challenge for forecasting and anomaly detection, LLMs hold considerable promise in addressing this issue. Through advanced preprocessing, intelligent feature extraction, and the integration of external knowledge, LLMs can unlock valuable insights hidden within unstructured text. Continued advancements in model development and training methodologies will be vital in overcoming the obstacles posed by noise and unstructured data, enabling more accurate and insightful predictive analyses.

\section{Datasets}\label{sec:datasets}

In the realm of forecasting and anomaly detection research, the availability of high-quality datasets is a critical factor for advancement. These datasets facilitate rapid development and fine-tuning of effective detection algorithms while also setting benchmarks for evaluating methodological performance. However, the acquisition of such datasets often entails significant financial, material, and workforce investments. The field is currently experiencing early development stages, characterized by challenges such as limited data quantity, complex sample characteristics, and missing labels, both essential for developing effective approaches. This section highlights prominent datasets utilized in LLM for forecasting and anomaly detection, which have been contributed by recent studies. An assessment of these datasets is conducted, pinpointing prevailing limitations and challenges in dataset generation, with the objective of guiding the creation of future datasets in this domain.

\subsection{Forecasting}

In the field of forecasting, the attributes of datasets hold paramount importance in determining the success and accuracy of predictive models. Essential characteristics include temporal resolution and range, where the granularity of time intervals and the overall time span covered by the dataset are critical for capturing the necessary details and trends. Completeness and continuity are equally important; datasets should be devoid of gaps and missing values to avoid inaccuracies and the need for complex imputation techniques. Variability and diversity within the data ensure the model is exposed to various scenarios, thus enhancing its ability to generalize and perform under varying conditions. The presence of non-stationary elements, which cause statistical properties to change over time, poses significant challenges and must be carefully considered and addressed. Seasonality and cyclic patterns are also crucial, as datasets must capture these recurring behaviors for models to forecast periodic fluctuations accurately. We have found the following datasets utilized in recent research of LLM for forecasting:

\bulletparagraph{Amazon Review}

The Amazon Review dataset \cite{ni2019justifying_recommendations} is a collection of reviews from Amazon.com. The dataset contains user reviews on Amazon shopping website from 2014-01-04 to 2016-10-02, with each review consisting of a product ID, reviewer ID, rating, and text. This dataset is used for time series rating forecasting and can be found in paper \cite{shi2023language_models}.

\bulletparagraph{Darts}

Darts \cite{herzen2022darts_user-friendly} is a Python library designed for easy manipulation, forecasting, and anomaly detection on time series data. Darts contains popular time series datasets for quick and reproducible experiments with A collection of 8 real univariate time series datasets. This dataset found application in the evaluation setup of works \cite{gruver2023large_language}.

\bulletparagraph{Electricity Consumption Load (ECL)}

The ECL dataset \cite{trindade2015electricityloaddiagrams20112014} from UCI collected in 2011 includes the electricity consumption values (in Kwh) of 321 users and 370 points per client. The dataset ensured that it contained no missing values. The analysis conducted in paper \cite{cao2024tempo_prompt-based, xue2023promptcast_a, jin2024time-llm_time, dong2023simmtm_a} was significantly based on this dataset.

\bulletparagraph{Integrated Crisis Early Warning System (ICEWS)}

The ICEWS dataset \cite{boschee2015icews_coded} is a collection of events extracted from news articles and other sources. The dataset contains 4.5 million events from 1995 to 2014, with each event consisting of a source, target, and type. These data consist of coded interactions between socio-political actors (i.e., cooperative or hostile actions between individuals, groups, sectors and nation states). Events are automatically identified and extracted from news articles. The research outlined in \cite{shi2023language_models} employed this dataset for its analysis.

\bulletparagraph{Informer / ETT / ETDataset}

The ETT or ETDataset proposed in the Informer paper \cite{zhou2021informer_beyond}, includes data from 69 transformer stations at 39 locations, covering aspects such as load, oil temperature, location, climate, and demand. This dataset is designed to support investigations into long sequence forecasting problems and includes subsets like ETTh1, ETTh2 for 1-hour-level data, and ETTm1 for 15-minute-level data. Each data point in the ETT dataset consists of the target oil temperature value and six power load features, with the data split into training, validation, and test sets. This dataset is commonly used for long-term forecasting and can be found in paper \cite{gruver2023large_language, zhou2023one_fits, cao2024tempo_prompt-based, jin2024time-llm_time, dong2023simmtm_a}.

\bulletparagraph{M3}

The M3-Competition dataset \cite{makridakis2000the_m3-competition} is a collection of time series data used in the M3-Competition, which is the third iteration of the M-Competitions. The M3-Competition dataset contains 3003 time series, selected to include various types of data (micro, industry, macro, etc.) and different time intervals. The time series in the dataset are either annual, quarterly, or monthly, and the number of observations for each series ranges between 14 and 126 observations. All values in the dataset are positive. This dataset constituted the core empirical basis for the investigation in paper \cite{jin2024time-llm_time}.

\bulletparagraph{M4}

The M4 dataset \cite{makridakis2020the_m4} is a collection of 100,000 time series used for the M4 competition. The dataset consists of a time series of yearly, quarterly, monthly, and other frequencies (weekly, daily, and hourly) data, which are divided into training and test sets. The minimum number of observations in the training test is 13 for yearly, 16 for quarterly, 42 for monthly, 80 for weekly, 93 for daily, and 700 for hourly series. The participants were asked to produce the following numbers of forecasts beyond the available data: six for yearly, eight for quarterly, 18 for monthly series, 13 for weekly series, and 14 and 48 forecasts, respectively, for the daily and hourly ones. This dataset played a crucial role in the research outcomes presented in paper \cite{jin2024time-llm_time}.

\bulletparagraph{Monash}

The Monash \cite{godahewa2021monash_time} forecasting archive contains 20 publicly available time series datasets from varied domains. The utilization of this dataset is documented in paper \cite{gruver2023large_language}.

\bulletparagraph{Text for Time Series (TETS)}

The TETS benchmark dataset was proposed and used in \cite{cao2024tempo_prompt-based} for short-term forecasting experiments. It is built upon the S\&P 500 dataset, combining contextual information and time series.

\subsection{Anomaly Detection}

In the realm of anomaly detection, the attributes of datasets are critical in shaping the efficacy and reliability of detection models. Anomaly detection tasks hinge on the ability to identify deviations from normal patterns, thus necessitating meticulously curated datasets to capture these nuances. One of the primary attributes of such datasets is the representation of ordinary versus anomalous data. The datasets must include a sufficient representation of normal data to establish a typical behavior baseline. Equally important is the inclusion of a diverse range of anomalies. These anomalies should vary in terms of their nature, intensity, and duration to ensure that the detection models can identify a broad spectrum of deviations. The balance between normal and anomalous data is also a critical factor. Typically, anomalies are rare occurrences in real-world scenarios, and this rarity needs to be reflected in the datasets. However, having too few anomalies can hinder the model's ability to learn to detect them effectively. Thus, a delicate balance must be struck to create a realistic and useful dataset. Another crucial aspect is the contextual richness of the datasets. Anomalies often make sense only within a specific context, and datasets need to provide sufficient contextual information. This includes temporal context, which can be crucial for identifying time-based anomalies, and other domain-specific information that helps understand the significance of the data points. The quality and cleanliness of the data are also paramount. Anomaly detection models can be sensitive to noise and errors in the data. High-quality datasets with minimal noise and errors are essential for developing robust models. Additionally, the presence of labeled anomalies, which have been accurately identified and categorized, can significantly aid in the training and evaluating detection models. In recent studies on LLM for anomaly detection, the following datasets have been identified as commonly employed:

\bulletparagraph{Blue Gene/L (BGL)}

BGL \cite{oliner2007what_supercomputers} is an open dataset containing 4,747,963 logs collected from a BlueGene/L supercomputer system consisting of 131,072 processors and 32,768GB of memory and was deployed at Lawrence Livermore National Labs in Livermore, California. The log contains alert and non-alert messages identified by alert category tags. Each log in the BGL dataset was manually labeled as either normal or anomalous. Out of the total, 348,460 log messages, which represent 7.34\% of the dataset, were identified as anomalous. The analysis conducted in paper \cite{chen2022bert-log_anomaly, lee2023lanobert_system, zhang2023logprompt_a, huang2023improving_log-based, shao2022log_anomaly, he2023parameter-efficient_log, almodovar2024logfit_log, le2021log-based_anomaly, huang2020hitanomaly_hierarchical} was significantly based on this dataset.

\bulletparagraph{Hadoop Distributed File System (HDFS)}

The HDFS dataset \cite{xu2009detecting_large-scale} is collected from more than 200 Amazon EC2 nodes. It consists of 11,175,629 log events, each associated with a block ID\@. These log messages form different log windows according to their block ID, reflecting a program execution in the HDFS system. For each execution, labels are provided to indicate whether anomalies exist. This dataset has 16,838 blocks of logs (2.93\%) indicating system anomalies. The analysis conducted in paper \cite{chen2022bert-log_anomaly, lee2023lanobert_system, zhang2023logprompt_a, huang2023improving_log-based, shao2022log_anomaly, he2023parameter-efficient_log, hu2023research_on, almodovar2024logfit_log, zhang2022logst_log, le2021log-based_anomaly, huang2020hitanomaly_hierarchical} was significantly based on this dataset.

\bulletparagraph{OpenStack}

The OpenStack log datasets from CloudLab \cite{du2017deeplog_anomaly} contain 1,335,318 log entries. Both normal logs and abnormal cases with failure injection are provided in this dataset. This dataset was crucial to the research outcomes presented in paper \cite{ott2021robust_and, hu2023research_on}.

\bulletparagraph{Spirit}

Spirit dataset \cite{oliner2007what_supercomputers} aggregates system log data from the Spirit supercomputing system at Sandia National Labs. There are more than 272 million log messages in total, of which more than 172 million log messages are labeled as anomalous on the Spirit dataset. The empirical evidence in \cite{le2021log-based_anomaly} was derived using this particular dataset.

\bulletparagraph{Server Machine Dataset (SMD)}

SMD \cite{su2019robust_anomaly} is a 5-week-long dataset collected from 28 server machines at a large Internet company. It includes data from 38 different sensors or metrics per machine, which monitor various aspects of the server's operation, such as CPU load, network usage, and memory usage. The data was recorded at 1-minute intervals, and domain experts have labeled anomalies and their anomalous dimensions in the SMD testing set. The research outlined in \cite{li2022evaluating_bert} employed this dataset for its analysis.

\bulletparagraph{Thunderbird}

The Thunderbird dataset \cite{oliner2007what_supercomputers} is an open dataset of logs collected from a Thunderbird supercomputer at Sandia National Labs. There are around 211 million log messages, and the log data contains normal and abnormal messages that are manually identified. This dataset played a crucial role in the research outcomes presented in paper \cite{lee2023lanobert_system, shao2022log_anomaly, he2023parameter-efficient_log, almodovar2024logfit_log, le2021log-based_anomaly}.

\bulletparagraph{Yahoo S5}

The Yahoo S5 dataset is a labeled open dataset for anomaly detection released by Yahoo Lab. Part of the time series is synthetic (i.e., simulated). In contrast, the other part comes from the real traffic of Yahoo services. The anomaly points in the simulated curves are algorithmically generated, and those in the real-traffic curves are labeled by editors manually. The dataset was prominently featured in the experimental findings of paper \cite{dang2021ts-bert_time}.

\subsection{Summary}

In forecasting, the emphasis on temporal resolution and range, completeness, continuity, variability, diversity, non-stationarity, and the presence of cyclic patterns and seasonality are fundamental for the success of predictive models. These attributes ensure that the datasets are reflective of real-world complexities and variations, enabling the models to capture and predict trends and fluctuations accurately. The datasets identified in recent research have been tailored to address these needs, although challenges in data acquisition, quality, and representation persist.

For anomaly detection, the focus is on the representation of normal versus anomalous data, the diversity of anomalies, the balance between normal and anomalous instances, contextual richness, and data quality. These factors are crucial in crafting datasets that accurately reflect real-world scenarios and enable LLMs to identify and distinguish between normal and anomalous behaviors effectively. The challenge lies in assembling datasets that are both realistic in their rarity of anomalies and rich in contextual detail to facilitate effective learning and detection.

Both fields face common challenges in dataset generation, including the need for large-scale, high-quality data that accurately captures the complexities of real-world scenarios. The issues of missing labels, noise in data, and the balance between various data characteristics are ongoing concerns. Future dataset creation in these domains should focus on addressing these challenges, ensuring greater accuracy and efficacy in forecasting and anomaly detection tasks. This will not only enhance the performance of current models but also pave the way for new advancements in the field.
\section{Evaluation Metrics}\label{sec:evaluation-metrics}

Evaluation metrics are indispensable tools for evaluating and comparing models in machine learning and statistical analysis, especially in domains such as forecasting and anomaly detection. In these areas, the ability of a model to predict future values based on historical data or identify irregular patterns that deviate from the norm is critical. Metrics in these contexts serve as quantitative indicators of a model's performance, offering insights into its predictive accuracy, reliability, and robustness under various conditions.

\subsection{Definition}

For forecasting, metrics such as Mean Absolute Error (MAE), Mean Squared Error (MSE), and Root Mean Squared Error (RMSE) are commonly employed to measure the deviation of predicted values from actual values, providing a clear picture of prediction accuracy. Additionally, the Mean Absolute Percentage Error (MAPE) and Symmetric Mean Absolute Percentage Error (sMAPE) offer insights into the relative prediction errors, making them particularly useful for comparing models across different scales or datasets. In this context, we have the following definitions:

\begin{itemize}
  \item \(\mathcal{N}\): the number of forecasting data points
  \item \(n\): \( n \in \{1, \dots, \mathcal{N}\} \)
  \item \(\mathcal{Y}_n\): the \(n\)-th ground truth
  \item \(\hat{\mathcal{Y}}_n\): the \(n\)-th forecasting value
\end{itemize}

In the realm of anomaly detection, the focus shifts towards identifying outliers effectively. Precision, Recall, and the F1 Score became crucial, quantifying the model's ability to correctly identify anomalies (true positives) while minimizing false alarms (false positives) and missed detections (false negatives). The Area Under the Receiver Operating Characteristic (AUROC) further provide comprehensive measures of a model's discriminative ability, balancing the trade-off between true positive rates and false positive rates across different threshold settings. In this given scope, the definitions are as follows:

\begin{itemize}
  \item \textbf{True Positive (TP)}: the total number of data samples that are correctly identified to be positive. This refers to the number of anomalies (or outliers) that the system correctly identifies as anomalies. Essentially, these are the instances where the system correctly detects an abnormal behavior or pattern that deviates from what's expected or normal.
  \item \textbf{True Negative (TN)}: the total number of data samples that are correctly identified to be negative. This refers to the number of normal instances that the system correctly identifies as normal. In other words, these are the cases where the system accurately recognizes that there is no anomaly present, and the behavior or pattern is as expected.
  \item \textbf{False Positive (FP)}: the total number of data samples that are incorrectly identified to be positive. This occurs when the system incorrectly identifies a normal instance as an anomaly. False positives are essentially false alarms, where the system flags normal behavior or data as being abnormal or suspicious when it is not. This can lead to unnecessary investigations or actions.
  \item \textbf{False Negative (FN)}: the total number of data samples that are incorrectly identified to be negative. This occurs when the system fails to identify an actual anomaly as an anomaly. In these cases, the system incorrectly considers abnormal behavior or patterns to be normal, potentially missing important or critical incidents.
\end{itemize}

These terms are crucial for evaluating the accuracy and effectiveness of applying LLM for anomaly detection. A high number of false positives might lead to wasted resources and desensitization to alerts, whereas a high number of false negatives could mean missing critical issues or breaches. Balancing sensitivity (minimizing FNs) and specificity (minimizing FPs) is critical to designing an effective anomaly detection system.

\subsection{Metrics}

In the pursuit of advancing the effectiveness and precision of LLMs in forecasting and anomaly detection, it is imperative to employ robust metrics that accurately capture the models' performance. This subsection delves into the diverse array of metrics that are instrumental in evaluating the outcomes of LLMs within these domains. Forecasting metrics offer unique insights into the model's predictive accuracy and reliability, while anomaly detection metrics provide a multi-dimensional view of model efficacy, balancing the detection accuracy with the rate of false alarms. This systematic exploration underscores the importance of choosing appropriate evaluation metrics and highlights how these metrics can guide the development and refinement of LLMs for enhanced performance in forecasting and anomaly detection tasks.

\subsubsection{Forecasting}

In the context of forecasting, a diverse array of metrics is employed to evaluate the accuracy and efficacy of predictive models meticulously. These metrics, each with its unique focus and application, serve as critical tools for quantitatively assessing how well a model's predictions align with actual outcomes. From the MAE, which provides a straightforward measure of average error magnitude, to the MAPE and its symmetric counterpart sMAPE, which offers insights into relative prediction errors, these metrics cater to various aspects of forecasting accuracy. The MSE and its derivative, the RMSE, emphasize the penalization of larger errors, making them especially pertinent in contexts where such errors are less tolerable. Additionally, the Root Mean Squared Percentage Error (RMSPE) and Mean Absolute Scaled Error (MASE) introduce normalized error measurements that facilitate model comparison across different scales or series. The Mean Absolute Ranged Relative Error (MARRE) and Overall Percentage Error (OPE) extend the toolkit by providing further nuances in error evaluation. Moreover, the Root Mean Squared Log Error (RMSLE) addresses handling asymmetric error distribution, which is particularly useful in skewed datasets. Lastly, the Overall Weighted Average (OWA) integrates multiple accuracy metrics into a single composite score, offering a holistic view of model performance. Collectively, these metrics equip forecasters with a comprehensive framework to scrutinize, compare, and enhance the predictive capabilities of their models, ensuring more informed decision-making and strategy development in various domains.

\bulletparagraph{Mean Absolute Error (MAE)}

MAE quantifies the average magnitude of errors in a collection of predictions, disregarding the errors' direction. It represents the mean of the absolute discrepancies between predicted values and actual observations across a dataset, treating all deviations with uniform importance.

\[
  \mathrm{MAE}(\mathcal{Y}_n, \hat{\mathcal{Y}}_n)=\frac{1}{\mathcal{N}} \sum_{n=1}^{\mathcal{N}}\left|\hat{\mathcal{Y}}_n-\mathcal{Y}_n\right|.
\]

Recent works \cite{gruver2023large_language, zhou2023one_fits, cao2024tempo_prompt-based, xue2023promptcast_a, xue2022leveraging_language, jin2024time-llm_time, jin2021trafficbert_pre-trained, dong2023simmtm_a} employed this particular metric for its evaluative procedures.

\bulletparagraph{Mean Absolute Percentage Error (MAPE)}

MAPE quantifies the precision of forecasts by representing the error as a percentage of the total. It's calculated as the average of the absolute percentage errors of the predictions. This characteristic makes MAPE very easy to interpret but can also be misleading if dealing with values close to zero.

\[
  \mathrm{MAPE}(\mathcal{Y}_n, \hat{\mathcal{Y}}_n)=\frac{100 \%}{\mathcal{N}} \sum_{n=1}^{\mathcal{N}}\left|\frac{\hat{\mathcal{Y}}_n-\mathcal{Y}_n}{\mathcal{Y}_n}\right|.
\]

The research detailed in \cite{zhou2023one_fits, jin2024time-llm_time, jin2021trafficbert_pre-trained} incorporated this metric in its assessment.

\bulletparagraph{Symmetric Mean Absolute Percentage Error (sMAPE)}

sMAPE is a variation of MAPE that is symmetric, meaning it treats over-forecast and under-forecast equally. It's considered more accurate than MAPE by some because it normalizes errors by the sum of the forecast and actual values, thus avoiding the issue of division by a small number.

\[
  \mathrm{sMAPE}(\mathcal{Y}_n, \hat{\mathcal{Y}}_n)=\frac{100 \%}{\mathcal{N}} \sum_{n=1}^{\mathcal{N}} \frac{2\left|\hat{\mathcal{Y}}_n-\mathcal{Y}_n\right|}{\left|\hat{\mathcal{Y}}_n\right|+\left|\mathcal{Y}_n\right|}.
\]

The metric discussed was applied in the analysis presented in \cite{zhou2023one_fits, cao2024tempo_prompt-based, jin2024time-llm_time}.

\bulletparagraph{Mean Squared Error (MSE)}

MSE calculates the mean of the squared discrepancies between predicted and true values, offering a measure of the average error magnitude. MSE gives more weight to more significant errors due to the squaring of each term, which can be particularly useful in some contexts where larger errors are more undesirable than smaller ones.

\[
  \mathrm{MSE}(\mathcal{Y}_n, \hat{\mathcal{Y}}_n)=\frac{1}{\mathcal{N}} \sum_{n=1}^{\mathcal{N}}{\left(\hat{\mathcal{Y}}_n-\mathcal{Y}_n\right)}^2.
\]

The application of this metric is elaborated in the evaluation section of \cite{zhou2023one_fits, cao2024tempo_prompt-based, jin2024time-llm_time, li2022evaluating_bert, sheng2023an_augmentable, dong2023simmtm_a}.

\bulletparagraph{Root Mean Squared Error (RMSE)}

RMSE is the square root of the mean squared error. It's a measure of the magnitude of the difference between the predictions of a model and the observed values. By taking the square root of MSE, RMSE converts the units back to the original output units, making interpretation easier.

\[
  \mathrm{RMSE}(\mathcal{Y}_n, \hat{\mathcal{Y}}_n)=\sqrt{\frac{1}{\mathcal{N}} \sum_{n=1}^{\mathcal{N}}{\left(\hat{\mathcal{Y}}_n-\mathcal{Y}_n\right)}^2}.
\]

Paper \cite{shi2023language_models, xue2023promptcast_a, xue2022leveraging_language, jin2021trafficbert_pre-trained, sheng2023an_augmentable} features the use of this metric in its experimental validation phase.

\bulletparagraph{Root Mean Squared Percentage Error (RMSPE)}

RMSPE is a normalized metric that expresses the average of the squares of the percentage errors between actual and forecasted values. It is beneficial for comparing forecasting errors across different data sets because it is scale-independent. The RMSPE is especially insightful when inspecting the error in terms of the percentage of the actual values, providing a clear picture of the relative size of the errors.

\[
  \mathrm{RMSPE}(\mathcal{Y}_n, \hat{\mathcal{Y}}_n)=\sqrt{\frac{1}{\mathcal{N}} \sum_{n=1}^{\mathcal{N}}{\left(\frac{\hat{\mathcal{Y}}_n-\mathcal{Y}_n}{\mathcal{Y}_n}\right)}^2}.
\]

\bulletparagraph{Mean Absolute Scaled Error (MASE)}

MASE measures the accuracy of forecasts relative to a naive benchmark prediction, typically the naive forecast from the previous period. This scaling makes MASE an excellent tool for comparing the performance of forecasting models across different data sets with varying scales. MASE is particularly advantageous since it is easy to interpret and does not require the forecast errors to be normally distributed.

\[
  \mathrm{MASE}(\mathcal{Y}_n, \hat{\mathcal{Y}}_n)= \frac{\frac{1}{\mathcal{N}} \sum_{n=1}^{\mathcal{N}} |\hat{\mathcal{Y}}_n - \mathcal{Y}_n|}{\frac{1}{\mathcal{N}-1} \sum_{n=2}^{\mathcal{N}} |\mathcal{Y}_n - \mathcal{Y}_{n-1}|}.
\]

As delineated in \cite{jin2024time-llm_time, jin2021trafficbert_pre-trained}, the metric was critical to their evaluative strategy.

\bulletparagraph{Mean Absolute Ranged Relative Error (MARRE)}

MARRE is a metric that assesses the absolute errors in relation to a specific range of the dataset, making it particularly useful for datasets where the range of data points is significant. MARRE helps in understanding the magnitude of errors in the context of the overall variation in the dataset.

\[
  \mathrm{MARRE}(\mathcal{Y}_n, \hat{\mathcal{Y}}_n)=\frac{1}{\mathcal{N}} \sum_{n=1}^{\mathcal{N}} \left( \frac{|\hat{\mathcal{Y}}_n - \mathcal{Y}_n|}{\max(\mathcal{Y}) - \min(\mathcal{Y})} \right).
\]

\bulletparagraph{Overall Percentage Error (OPE)}

OPE aggregates the total errors as a percentage of the total actual values. It provides a single, comprehensive figure that reflects the overall accuracy of the forecasts in relation to the actual observations, offering a macroscopic view of forecasting performance.

\[
  \mathrm{OPE}(\mathcal{Y}_n, \hat{\mathcal{Y}}_n)=\frac{\sum_{n=1}^{\mathcal{N}} |\hat{\mathcal{Y}}_n - \mathcal{Y}_n|}{\sum_{n=1}^{\mathcal{N}} \mathcal{Y}_n} \times 100\%.
\]

\bulletparagraph{Root Mean Squared Log Error (RMSLE)}

RMSLE is used to measure the ratio between actual and predicted values. By taking the log of the predictions and actual values before calculating the mean square error, RMSLE reduces the impact of significant errors and is less sensitive to outliers than RMSE\@. It's particularly useful when you don't want to penalize enormous differences when both the actual and predicted values are big numbers.

\[
  \mathrm{RMSLE}(\mathcal{Y}_n, \hat{\mathcal{Y}}_n)=\sqrt{\frac{1}{\mathcal{N}} \sum_{n=1}^{\mathcal{N}} {(\log(\hat{\mathcal{Y}}_n + 1) - \log(\mathcal{Y}_n + 1))}^2}.
\]

\bulletparagraph{Overall Weighted Average (OWA)}

OWA is a specific metric that was introduced as part of the M4 forecasting competition \cite{makridakis2020the_m4}, which aimed to advance the field of forecasting by comparing and evaluating the performance of various forecasting models across multiple time series datasets. OWA is particularly notable because it combines aspects of both accuracy and scalability into a single metric, making it a comprehensive measure for evaluating forecasting models.

OWA is calculated by averaging two key components: the MASE and the sMAPE\@. These two metrics are chosen because they provide complementary perspectives on forecasting performance: MASE offers a scale-independent measure of error relative to a simple naive benchmark, and sMAPE provides a percentage-based measure of error that is symmetric, treating over-forecasts and under-forecasts equally.

\[
  \mathrm{OWA}(\mathcal{Y}_n, \hat{\mathcal{Y}}_n)=\frac{1}{2} \left( \frac{\mathrm{MASE}}{\mathrm{MASE}_{\mathrm{Naive2}}} + \frac{\mathrm{sMAPE}}{\mathrm{sMAPE}_{\mathrm{Naive2}}} \right).
\]

In this context, \(\mathrm{MASE}_{\mathrm{Naive2}}\) and \(\mathrm{sMAPE}_{\mathrm{Naive2}}\) refer to the MASE and sMAPE scores obtained by a naive forecasting method (Naive2), typically a seasonal naive method that uses the last observed value of the same season as the forecast. This normalization against a naive benchmark allows OWA to reflect both the absolute and relative improvement of a forecasting method over a simple but commonly applicable baseline.

Paper \cite{jin2024time-llm_time} features the use of this metric in its experimental validation phase.

\subsubsection{Anomaly Detection}

In the field of anomaly detection, the effectiveness of a model is significantly determined by its ability to identify outliers and minimize missed detections and false alarms accurately. The key metrics used to evaluate such models include Accuracy, Precision, Recall, True Negative Rate (TNR), False Positive Rate (FPR), False Negative Rate (FNR), the F1 Score, and the AUROC\@. These metrics, derived from the fundamental concepts of TP, TN, FP, and FN, provide a comprehensive framework for assessing the performance of anomaly detection systems.

\bulletparagraph{Accuracy}

Accuracy quantifies the fraction of correct predictions, encompassing both true positives and true negatives, relative to the overall sample size evaluated. Accuracy is the most straightforward and intuitive performance measure, giving a general idea of how often the model is correct. While accuracy is straightforward, it may not always be the best metric for anomaly detection, especially in datasets where anomalies are rare (imbalanced datasets). In such cases, a model might achieve high accuracy by predicting the majority class (normal) most of the time while failing to detect many anomalies.

\[
  \mathrm{Accuracy} = \frac{TP + TN}{TP + TN + FP + FN}.
\]

The research detailed in \cite{li2022evaluating_bert, zhang2023logprompt_a, shao2022log_anomaly} incorporated this metric in its assessment.

\bulletparagraph{Precision}

Precision, also known as Positive Predictive Value, quantifies the number of correct positive identifications made out of all positive identifications (correct and incorrect). Precision is crucial in scenarios where the cost of false positives is high. For instance, in transaction anomaly detection, a false positive (flagging a legitimate transaction as fraudulent) could inconvenience customers and erode trust. High precision indicates that when the model predicts an anomaly, it is likely to be a true anomaly.

\[
  \mathrm{Precision} = \frac{TP}{TP + FP}.
\]

The application of this metric is elaborated in the evaluation section of \cite{ott2021robust_and, li2022evaluating_bert, jin2022symlm, zhang2023logprompt_a, dang2021ts-bert_time, huang2023improving_log-based, gupta2023learning_representations, karlsen2023exploring_semantic, hu2023research_on, almodovar2024logfit_log, zhang2022logst_log, le2021log-based_anomaly, huang2020hitanomaly_hierarchical}.

\bulletparagraph{Recall / True Positive Rate (TPR)}

Recall, also known as Sensitivity or True Positive Rate (TPR), measures the proportion of actual positives correctly identified, emphasizing the model's ability to capture all relevant positive outcomes. In the context of anomaly detection, a high recall means that the model is effective at catching anomalies, which is critical in situations where missing an anomaly can have severe consequences, such as in predictive maintenance or health monitoring.

\[
  \mathrm{Recall\ (TPR)} = \frac{TP}{TP + FN}.
\]

The research detailed in \cite{ott2021robust_and, li2022evaluating_bert, zhang2023logprompt_a, dang2021ts-bert_time, huang2023improving_log-based, gupta2023learning_representations, shao2022log_anomaly, jin2022understanding, he2023parameter-efficient_log, karlsen2023exploring_semantic, hu2023research_on, almodovar2024logfit_log, zhang2022logst_log, le2021log-based_anomaly, huang2020hitanomaly_hierarchical} incorporated this metric in its assessment.

\bulletparagraph{True Negative Rate (TNR)}

TNR, also known as Specificity, quantifies the proportion of actual negatives that are correctly identified, reflecting the model's ability to identify negative outcomes. High TNR means few normal instances are incorrectly flagged as anomalies, which helps reduce false alarms and maintain trust in the system's predictions.

\[
  \mathrm{TNR} = \frac{TN}{TN + FP}.
\]

\bulletparagraph{False Positive Rate (FPR)}

FPR measures the proportion of false positives out of the total actual negatives, indicating how often false alarms occur. It is the rate at which regular instances are wrongly classified as anomalies. In many applications, it's crucial to minimize FPR to avoid the costs associated with false alarms, such as wasted resources or unnecessary anxiety.

\[
  \mathrm{FPR} = \frac{FP}{FP + TN}.
\]

\bulletparagraph{False Negative Rate (FNR)}

FNR quantifies the proportion of false negatives out of the total actual positives, indicating the model's miss rate. It quantifies the model's failure to detect anomalies. A high FNR indicates that many anomalies go undetected, potentially leading to missed opportunities for intervention in critical situations.

\[
  \mathrm{FNR} = \frac{FN}{TP + FN}.
\]

\bulletparagraph{F1 Score}

The F1 Score is the harmonic mean of precision and recall, providing a single score that balances the trade-off between precision and recall. It is a useful metric for evaluating the overall performance of a model, especially when there is an uneven class distribution (e.g., a large number of normal instances and a small number of anomalies). It is particularly valuable in imbalanced datasets, where TPs are much less common than TNs. Therefore, F1 Score is particularly useful in anomaly detection because it balances the trade-off between minimizing false alarms (FP) and minimizing missed detections (FN).

\[
  \mathrm{F_1} = 2 \cdot \frac{\mathrm{Precision} \cdot \mathrm{Recall}}{\mathrm{Precision} + \mathrm{Recall}}.
\]

This metric was a key component in the experimental design of \cite{chen2022bert-log_anomaly, lee2023lanobert_system, dang2020time_series, ott2021robust_and, li2022evaluating_bert, zhang2023logprompt_a, dang2021ts-bert_time, huang2023improving_log-based, gupta2023learning_representations, jin2023understand, shao2022log_anomaly, he2023parameter-efficient_log, karlsen2023exploring_semantic, hu2023research_on, almodovar2024logfit_log, zhang2022logst_log, le2021log-based_anomaly, huang2020hitanomaly_hierarchical}.

\bulletparagraph{Area Under the Receiver Operating Characteristic (AUROC)}

AUROC represents the likelihood of the model distinguishing between the positive class (anomalies) and the negative class (normal cases) \cite{hanley1982the_meaning}. It reflects the model's ability to classify outcomes correctly at various threshold levels, providing a comprehensive measure of performance across all possible classification thresholds. AUROC stands out as a comprehensive measure that evaluates a model's ability to distinguish between classes across all thresholds. The ROC curve plots the TPR against the FPR at various threshold settings. AUROC represents the probability that a model will rank a randomly chosen positive instance higher than a randomly chosen negative one. A model with an AUROC of 1.0 is perfect, distinguishing between all positive and negative instances correctly, while a score of 0.5 suggests no discriminative ability, equivalent to random guessing.

\[
  \mathrm{AUROC}=\int_0^1 \frac{TP}{TP+FP} \mathrm{~d} \frac{FP}{FP+TN}.
\]

AUROC is particularly informative in anomaly detection because it provides insight into the model's performance across a range of conditions, allowing for the evaluation of the model's generalizability and robustness. It helps identify the best model that manages the trade-off between detecting as many anomalies as possible (high TPR) while keeping false alarms (high FPR) to a minimum. This is crucial in real-world applications where the cost of false positives and false negatives can vary significantly, and choosing an operating point (a specific threshold) that balances these costs is essential.

As delineated in \cite{lee2023lanobert_system}, the metric was critical to their evaluative strategy.

\section{Forecasting with Large Language Models}\label{sec:forecasting}

In the domain of artificial intelligence, LLMs have emerged as pivotal instruments for advancing forecasting methodologies across a myriad of fields. This section delves into the transformative role these models play in predicting future events and trends with unprecedented accuracy. This section is meticulously structured to cover the versatile applications of LLMs in forecasting, starting with time series forecasting, a fundamental approach that is further delineated into short-term and long-term forecasting. Each of these subcategories showcases the specific challenges and solutions that LLMs address, highlighting their flexibility and efficiency. Moving beyond traditional time series analysis, the discussion extends to traffic flow forecasting, illustrating how LLMs enhance urban mobility and reduce congestion through predictive analytics. Furthermore, the section explores the profound impact of LLMs in healthcare clinical prediction, where they offer groundbreaking insights into patient outcomes, disease progression, and treatment efficacy. Through this comprehensive examination, we aim to underscore the significant advancements LLMs bring to forecasting practices, fostering a deeper understanding of their capabilities and potential for innovation in various sectors.

\subsection{Time Series Forecasting}

This section embarks on an in-depth exploration of how LLMs have revolutionized the analysis and prediction of sequential data over time. This critical area of forecasting serves as the backbone for numerous applications, ranging from financial market predictions to energy consumption planning. By dividing the discussion into short-term and long-term forecasting, this section meticulously addresses the nuances and specificities of forecasting at different horizons. Short-term forecasting focuses on the immediate future, where precision and speed are paramount, highlighting LLMs' ability to process and analyze data for near-term predictions rapidly. Conversely, long-term forecasting examines trends and patterns over extended periods, demonstrating how LLMs can identify underlying signals amidst noise, providing valuable foresight for strategic planning and decision-making.

Gruver \textit{et al.} (2023) \cite{gruver2023large_language} introduces an innovative approach for forecasting time series by utilizing LLMs like GPT-3 and LLaMA-2. This method involves encoding time series data as strings of numerical digits, thereby converting the forecasting challenge into predicting the next token in a sequence akin to text prediction. This strategy enables LLMs to extrapolate future values in time series data without any task-specific prior training. The effectiveness of this approach is noted to be on par with or superior to traditional time series models explicitly designed for such tasks. The authors emphasize the utility of LLMs in capturing the nuanced dynamics of time series forecasting due to their capability to encode multimodal distributions, which is advantageous for representing the inherent variability and repeated patterns found in many time series datasets. This attribute, combined with LLMs' inclination towards simplicity and pattern repetition, is critical to their success in time series analysis. One of the major advantages highlighted by the authors is the zero-shot nature of their approach, which prevents the need for detailed knowledge of model fine-tuning or the extensive computational resources typically required. This aspect is particularly beneficial when data is scarce, thus eliminating the necessity for extensive model training or fine-tuning. The broad generalization capacity of LLMs, thanks to their extensive pre-training, allows for effective pattern recognition and extrapolation without the need for domain-specific model development. Moreover, the methodology described facilitates handling missing data through non-numerical text, integrating textual information alongside numerical time series data, and explaining predictions by answering questions. This comprehensive capability demonstrates the versatility of LLMs in dealing with complex forecasting tasks. However, the authors also caution that larger LLMs, such as GPT-4, may not always yield improved performance over smaller counterparts like GPT-3. This is attributed to differences in number tokenization and a lack of reliable uncertainty calibration, potentially due to modifications in model training procedures like RLHF\@. This groundbreaking work underscores the potential of leveraging LLMs for time series forecasting, showcasing their adaptability across diverse domains and ability to simplify the forecasting process without compromising accuracy or requiring extensive domain expertise.

Zhou \textit{et al.} (2023) \cite{zhou2023one_fits} demonstrates the effectiveness of using pre-trained language and computer vision models for various time series analysis tasks without modifying their architecture. It's very challenging to efficiently utilize pre-trained models from other domains, like natural language processing and computer vision, for diverse time series analysis tasks, aiming to overcome the need for domain-specific architectural changes and to harness the power of these pre-trained models for improved performance in time series analysis. This work employs a novel architecture for time series analysis, using parameters from pre-trained NLP transformer models. Specifically, the study focuses on the GPT2 model and experiments with other models like BERT and BEiT. This approach represents a significant shift from traditional methods, as it leverages the strengths of pre-trained models from different domains (like language and vision) for time series analysis, thus exploring the universality and versatility of these models in a new context. The zero-shot performance of the proposed approach still lags behind the state-of-the-art methods. This suggests that while the method is effective in many scenarios, it may not yet be fully optimized for zero-shot learning tasks where the model makes predictions without any prior examples from the specific task domain. This work proposed a unified framework that uses a frozen pre-trained language model to achieve state-of-the-art or comparable performance in all major types of time series analysis tasks. This includes time series classification, short/long-term forecasting, imputation, anomaly detection, and few-shot and zero-shot forecasting, supported by thorough and extensive experiments. Theoretical and empirical findings show that self-attention in transformers performs a function similar to Principal Component Analysis (PCA), helping to explain the universality of transformer models. The authors demonstrated the universality of their approach by successfully applying a pre-trained transformer model from another backbone model (like BERT) or modality (such as computer vision) to power time series forecasting.

Shi \textit{et al.} (2023) \cite{shi2023language_models} investigates whether LLMs can reason about real-world events and improve event prediction. The motivation behind this objective is the potential usefulness of LLMs in handling event sequences that are often accompanied by rich text information. Large language models have shown impressive performance on various reasoning tasks, and the authors aim to explore their capabilities in reasoning about real-world events. Event sequences are often accompanied by text information, and LLMs excel at handling textual data. Therefore, integrating LLMs into event prediction models can potentially improve their performance. The authors propose a framework called LAMP incorporating a large language model in event prediction. They use abductive reasoning to suggest possible causes for event predictions and retrieve relevant events from history to support these predictions. One potential threat of this paper is the reliance on large language models, which may have limitations in terms of data leakage and generalization. However, the authors address these concerns by verifying the absence of data leakage and demonstrating the generalization capabilities of LLMs in their experiments. Another threat is the limited evaluation of specific datasets, which may not fully represent the complexity of real-world event prediction tasks. However, the authors mitigate this threat by conducting experiments on multiple datasets and demonstrating consistent improvements over baseline models. The proposed LAMP framework is innovative and practical, as it integrates a large language model into event prediction models, leveraging the reasoning capabilities of LLMs. The framework provides insightful empirical findings through extensive experiments on challenging real-world datasets, demonstrating its significant improvement over state-of-the-art event sequence models. This work presents a well-structured review of relevant literature, discussing the existing event sequence models and their limitations, as well as the potential of LLMs in event prediction. It addresses the threats of data leakage by verifying that the LLMs used in the experiments were trained on data that does not include the datasets used in the experiments.

Cao \textit{et al.} (2024) \cite{cao2024tempo_prompt-based} proposed TEMPO, which aims to leverage the strengths of transformer-based models for their ability and attention mechanism to handle sequential data, learn from context, and apply it to time-series forecasting tasks. It designed and evaluated an approach for time series forecasting using a method adapted from GPTs. Because of the success of GPTs in NLP, TEMPO hypothesizes that the same architecture can be adapted to understand and predict time series data, which is inherently sequential. It leverages the power of pre-training models and finds self-attention mechanisms to be good at capturing dependencies in sequential data. Inspired by the prompt-based GPTs, like ChatGPT, TEMPO uses historical data points as prompts, like a conversation. By leveraging the power of pre-training, TEMPO can generalize across different time series domains and tasks. TEMPO beats traditional methods in accuracy and other benchmarks in many datasets. However, it has computational cost and dataset requirements of quality and quantity concerns.

Xue \textit{et al.} (2023) \cite{xue2023promptcast_a} presented PromptCast, which aims to establish a new paradigm that transforms the traditional numerical time series forecasting task into a prompt-based task. This approach is motivated by the successes of pre-trained language foundation models in NLP\@. One of the primary challenges is the effective translation of numerical time series data into textual prompts that language models can process. This approach needs more benchmarks for evaluating the prompt-based methods and further evaluation under real-world scenarios, like financial market crashes.

Zarzà \textit{et al.} (2023) \cite{dezarza2023llm_multimodal} studies the efficacy of modern deep learning methods for forecasting traffic accidents and enhancing Level-4 and Level-5 autonomous driving assistants with actionable visual and language cues. The motivation is to improve city planning and public safety by predicting accidents using a rich dataset of accident occurrences, thus paving the way for safer and smarter cities driven by data-driven decision-making. The authors identify the growing problem of traffic congestion and accidents in urban centers and the need for predictive analytics to mitigate these issues. This work acknowledges that traditional statistical models may only partially capture the complex interplay of factors leading to traffic accidents. The authors propose the use of advanced deep learning methods, such as Transformers, in conjunction with traditional time series models like ARIMA and Prophet for improved accident forecasting. They introduce the novel idea of employing LLMs and Visual Language Models (VLMs) to provide real-time interventions in autonomous driving. The rationale includes an in-depth analysis of feature importance using principal component analysis to identify key factors contributing to accidents. The paper also explores the concept of multimodality by utilizing a visual language model (LLaVA) to bridge visual and linguistic cues for enhancing autonomous driving systems. However, this work may face challenges in demonstrating the real-world applicability and scalability of the proposed methods, especially in diverse urban environments. There may be concerns regarding the interpretability and transparency of the deep learning models, which are often considered ``black boxes.'' The reliance on a specific dataset for analysis could limit the generalizability of the findings to other regions or conditions not represented in the data. The integration of LLMs and VLMs into autonomous driving systems might raise questions about the safety and reliability of language-based interventions in real-time traffic situations. The paper's proposed methods may need to address the computational complexity and resource requirements associated with processing large multimodal datasets in real time. The work presents an innovative methodology that combines modern deep learning techniques with traditional time series models for traffic accident forecasting. It contributes to the field by introducing the use of compact LLMs, such as LLaMA-2 and Zephyr-7b-a, for real-time interventions in autonomous driving. The study provides empirical findings on feature importance using PCA loadings, which can inform the development of more effective predictive models. It offers a well-structured review of the relevant literature, situating the current work within the broader context of traffic safety and autonomous driving research. The introduction of LLaVA as a multimodal model that integrates visual and linguistic cues is a notable contribution, potentially enhancing the responsiveness of autonomous driving systems. The paper has practical implications for city planners, traffic management agencies, and emergency services by providing actionable insights for optimizing resource allocation and intervention strategies.

Xue \textit{et al.} (2022) \cite{xue2022leveraging_language} proposes a novel pipeline named  AuxMobLCast that leverages language foundation models to discover temporal sequential patterns in human mobility forecasting tasks. In the new pre-train and fine-tune paradigm, a foundation model is pre-trained with large-scale data and then adapted to solve various downstream tasks. However, this shift only appears in the NLP and CV fields. How to apply a foundation model for spatio-temporal forecasting and human mobility prediction still needs to be explored. In the time-series data forecasting domain, especially with the human mobility data, there has yet to be any existing work on directly using pre-trained language foundation models for human mobility prediction due to the sequential numerical data format. The authors denote a set of POIs, and each POI contains a history record of customer visits on N continuous days. The author formulated the human mobility forecasting problem as predicting the number of visits the next day, given the historical observation. Then, three types of mobility prompting are introduced to convert the sequential observation into language description to leverage pre-trained language models for forecasting human mobility. Finally, the paper proposes a novel pipeline, AuxMobLCast, based on the general encoder-decoder framework with an auxiliary classification task to classify the POI category. One limitation of this study is about the mobility prompt generation. In the future, they plan to thoroughly investigate mobility prompts based on the recent prompt learning techniques. An automatic approach for transforming diverse sequential numerical behavior data and various types of time-series data will be beneficial in exploring the forecasting ability of pre-trained language models. In addition, how to explore pre-trained language models for multi-variate time-series data forecasting could be another interesting future direction.

Jin \textit{et al.} (2024) \cite{jin2024time-llm_time} demonstrates that large language models' rich semantic understanding and contextual learning abilities can be effectively adapted for the structurally distinct challenge of time series forecasting. This work seeks to establish methods and techniques for this reprogramming process and to evaluate the efficacy of these adapted models in time series forecasting, potentially offering a new avenue for utilizing existing language models in diverse applications beyond text-based tasks. The motivation of this work is to harness the advanced capabilities of large language models for the task of time series forecasting, thereby expanding their applicability beyond traditional text-based tasks. The rationale of the paper is to demonstrate that the rich semantic understanding and contextual learning abilities of large language models can be effectively adapted for the structurally distinct challenge of time series forecasting. The work's limitation lies in its reliance on the inherent capabilities of pre-trained language models, which may not be ideally suited or optimized for the specific nuances and complexities of time series data. The authors use Llama2-7B as the foundation model for evaluating two public benchmarking datasets compared to the baseline models from open-source TSlib. The ETT dataset is used to assess the long-term forecasting capability, while the M4 dataset is employed for short-term forecasting. The proposed method outperforms other baseline models regarding MSA and MAE metrics.

Li \textit{et al.} (2022) \cite{li2022evaluating_bert} evaluates the performance of BERT, a prominent language model, in two distinct applications: cloud-edge time series forecasting and sentiment analysis, utilizing prompt learning techniques. Their study aims to assess BERT's effectiveness and limitations in these areas to understand its applicability and potential for improvement in such tasks. This work investigates the capability of BERT in cloud-edge time series forecasting, a task that requires logical reasoning and an understanding of temporal data trends. Given its primary design for language understanding, the challenge is determining how well BERT can perform in this context. The authors aim to provide insights into these challenges and BERT's applicability and limitations in addressing them. This work applies prompt learning with BERT for cloud-edge time series forecasting and sentiment analysis. This approach seeks to leverage BERT's language understanding capabilities by framing the forecasting and sentiment analysis tasks to align with natural language processing. Prompt learning, which involves creating prompts that guide the model to understand and execute specific tasks, is used to adapt BERT, initially designed for language tasks, to these new application areas. The effectiveness of this method is then evaluated to understand how well BERT can handle these challenges. The potential limitations of this paper include a limited scope, as the study may not encompass a wide range of scenarios or datasets, potentially affecting the generalizability of the findings. Additionally, the inherent limitations of BERT, particularly in non-language tasks, and the constraints of the methodological approach, such as prompt learning, might impact the results and their broader applicability. These factors suggest that while the study provides valuable insights, its conclusions might be specific to the contexts and models tested.

Sheng (2023) \cite{sheng2023an_augmentable} proposes a scheme to train models on multimodal data combined with external knowledge bases, fine-tune GPT-4 on domain-specific data, train models on multimodal data, and equip models with probabilistic reasoning capabilities to achieve analysis and Interpret financial and technical data to generate strategic insights and make future forecasts. The original datasets may pose a threat that financial data is easily manipulated, which could lead to inaccurate predictions. This work proposes a training scheme that combines multimodal data with external knowledge bases and domain-specific data, thereby improving the accuracy and reliability of domain-specific output from large language models.

Dong \textit{et al.} (2023) \cite{dong2023simmtm_a} introduces SimMTM, a streamlined pre-training framework for masked time-series modeling, aimed at enhancing the efficacy of time series analysis tasks like forecasting and classification. The framework's core strategy involves learning to reconstruct the original time series by leveraging multiple masked series. This initiative stems from the recognition that the most significant semantic information within time series is encapsulated in temporal variations, which pose annotation challenges due to their inherent complexity. The paper tackles the problem arising from conventional masked modeling techniques, where random masking of time points could obliterate critical temporal variations, complicating the task of reconstruction to the extent that it hampers effective representation learning. The proposed methodology is underpinned by the manifold perspective on masked modeling, positing that while direct reconstruction might be thwarted by the loss of crucial temporal variations, utilizing multiple neighbors (or multiple masked series) for reconstruction can mutually compensate for this loss. This facilitates a more manageable reconstruction process. Moreover, this technique implicitly conditions the model to discern the local manifold structure of the time series, thereby fostering more robust representation learning.

\subsection{Event Sequence Prediction}
Event sequences and time series data are two fundamental concepts within the realm of data analysis and predictive modeling, each serving unique purposes and offering distinct insights. An event sequence, by its very nature, comprises a series of discrete actions or occurrences, meticulously cataloged based on the sequence in which they transpire. Unlike time series data, which is inherently quantitative and often measured at regular intervals, event sequences emphasize the order and timing of events without necessarily adhering to a uniform time scale \cite{zuo2020transformer,zhang2020self,yang2022transformer}. This distinction is crucial as it underpins the divergent analytical approaches and methodologies applied to each data type. While time series analysis focuses on understanding trends, seasonality, and patterns over time, event sequence analysis delves into the intricacies of the relationships and dependencies between individual events. This analysis can uncover complex behavioral patterns and sequences of actions, which are particularly valuable in domains such as user behavior analysis, system logs, and transaction sequences, where the temporal ordering and occurrence of events hold significant analytical weight.

LLMs have become instrumental in event sequence prediction, offering diverse capabilities and applications. Two key aspects of LLMs in event prediction are highlighted below, with references to the relevant papers. One significant aspect of LLMs is their ability to revolutionize event prediction through advanced reasoning techniques. Xue \textit{et al.} (2023) \cite{xue2023prompt} delve into the transformative potential of LLMs in advancing event prediction tasks through abductive reasoning in a few-shot setting. This aspect explores the core strengths of LLMs, exemplified by models like GPT-3. LLMs excel in understanding the contextual nuances of events, capturing intricate long-term dependencies, and exhibiting impressive generalization capabilities. Their innate ability to contextualize and reason over diverse data sources empowers event prediction systems to provide more accurate and insightful forecasts. Nakshatr \textit{et al.} (2023) \cite{nakshatri2023using} proposed a generalized framework for newsflow clustering that automatically extracts potentially critical news events that attract high media attention by analyzing the temporal trends of news articles. Another critical dimension in event sequence prediction is the utilization of LLMs to handle streaming event sequences. Shi \textit{et al.} (2023) \cite{shi2023language_models} addresses the unique challenges posed by continuous streams of event data and presents innovative solutions enabled by LLMs. In real-world scenarios, event data often arrives in a continuous and dynamic stream, where the distribution of patterns may shift over time. Privacy concerns and memory constraints further complicate the task of continuous monitoring of event sequences. LLMs, with their adaptive and context-aware nature, offer a promising avenue for addressing these challenges.

\subsection{Traffic Flow Forecasting}

This section delves into the critical application of LLMs in addressing one of the most pressing challenges in urban planning and mobility management. This segment illuminates how LLMs are leveraged to predict traffic conditions, enabling cities and transportation authorities to optimize traffic flow, reduce congestion, and enhance road safety. By harnessing the power of vast datasets, including historical traffic patterns~\cite{ruan2022learning}, real-time road conditions~\cite{mo2022cvlight}, and socio-economic factors~\cite{szirmai2005dynamics}, LLMs offer unparalleled accuracy in forecasting traffic volumes and speeds across different times and locations. This predictive capability is pivotal for planning efficient public transportation schedules, designing intelligent traffic management systems, and facilitating emergency response strategies. The discussion in this section underscores the transformative potential of LLMs in shaping the future of urban mobility and transportation infrastructure.

Jin \textit{et al.} (2021) \cite{jin2021trafficbert_pre-trained} proposed TrafficBERT to address the challenge of accurately forecasting traffic flow over long ranges, which is a critical aspect of managing and optimizing traffic systems. Traditional traffic prediction models often struggle with capturing the intricate spatiotemporal dynamics of traffic flow. TrafficBERT, by leveraging the BERT model, aims to overcome these limitations. It is designed to better understand and predict complex traffic patterns, ultimately aiding in more efficient traffic management, reducing congestion, and enhancing road safety. The use of such advanced predictive models reflects the growing need for sophisticated tools in the realm of intelligent transportation systems. The objective of the TrafficBERT is to develop a model that can effectively forecast long-range traffic flow. By using a pre-trained BERT framework, TrafficBERT aims to analyze and predict traffic patterns and flow with high accuracy. This involves understanding and capturing the complex spatiotemporal correlations in traffic data, which is essential for accurate traffic forecasting over extended periods and across various road conditions. The rationale is that a model adept at understanding the nuanced patterns in language data can similarly excel in interpreting traffic patterns, thereby providing more accurate and reliable long-range traffic flow predictions. This approach aims to enhance traffic management and planning, reduce congestion, and improve road safety. The effectiveness of TrafficBERT hinges on the quality and diversity of training data, with potential risks of overfitting and limited adaptability to sudden changes in traffic conditions. Also, its complexity demands substantial computational resources and expertise, and raises privacy concerns regarding the use of traffic data, emphasizing the need for careful management and implementation in traffic systems. The contribution of TrafficBERT lies in its innovative approach to traffic flow forecasting, utilizing the BERT model to analyze and predict traffic patterns with high accuracy over long ranges. By adapting a model proven in natural language processing to the domain of traffic management, TrafficBERT demonstrates enhanced capability in understanding complex spatial and temporal dependencies in traffic data. This advancement represents a significant step forward in the field of intelligent transportation systems, offering a more sophisticated tool for traffic analysis and management.

\subsection{Healthcare Clinical Prediction}

This section delves into the transformative potential of LLMs within the healthcare sector, highlighting their role in advancing predictive analytics for patient care and clinical outcomes. In this critical exploration, we uncover how LLMs harness vast arrays of clinical data, including electronic health records, medical imaging \cite{xiao2022dual, zeng2022graph}, and genomic information, to forecast disease progression, patient outcomes, and treatment responses with remarkable precision. This section elucidates the complex methodologies LLMs employ to navigate the intricacies of medical data, offering insights into their ability to identify patterns and correlations that elude traditional analytical methods. By integrating these advanced predictive models, healthcare professionals can achieve a more nuanced understanding of patient health, enabling personalized treatment plans, early intervention strategies, and improved resource allocation. Through a detailed examination of LLMs' impact on healthcare clinical prediction, this segment aims to showcase the profound implications for patient care, medical research, and the broader healthcare ecosystem, underscoring the pivotal role of AI-driven innovations in shaping the future of medicine.

Jiang \textit{et al.} (2023) \cite{jiang2023health_system-scale} proposed an LLM-based system that can integrate in real-time with clinical workflows centered around writing notes and placing electronic orders, presenting the results from developing, evaluating, deploying, and prospectively assessing NYUTron. This approach relies on the fact that all clinically useful data and medical professionals' decision-making. The authors showed unstructured clinical notes from the electronic health record can enable the training of clinical language models, which can be used as all-purpose clinical predictive engines with low-resistance development and deployment. This approach leverages recent advances in natural language processing to train a large language model for medical language (NYUTron) and fine-tune it across a wide range of clinical and operational predictive tasks.

\section{Anomaly Detection using Large Language Models}\label{sec:anomaly-detection}

The advent of LLMs has significantly broadened the horizons of anomaly detection, offering sophisticated solutions to identify irregularities across diverse datasets and domains. This section embarks on a comprehensive examination of how LLMs are being utilized to pinpoint deviations that could signify errors, fraud, system failures, or cyber threats. This exploration begins with time series anomaly detection, where LLMs analyze sequential data to detect unusual patterns, benefiting industries reliant on continuous monitoring, such as finance, manufacturing, and energy. Moving forward, the discussion transitions to anomaly log analysis, highlighting the capacity of LLMs to sift through vast quantities of log data to identify and classify anomalies, thereby enhancing IT security and operational efficiency. The section on microservice anomaly detection showcases the application of LLMs in the increasingly complex domain of cloud computing and distributed systems, where they play a crucial role in maintaining system health and security by detecting anomalies at the microservice level. This detailed exploration aims to illuminate the cutting-edge methodologies and impactful applications of LLMs in anomaly detection, underscoring their critical role in safeguarding and optimizing modern digital infrastructures.

\subsection{Time Series Anomaly Detection}

This section delves into the intricate world of identifying outliers in sequential data sets, where LLMs have become invaluable tools. This critical facet of anomaly detection focuses on uncovering patterns that deviate from the norm within time-dependent data, a task essential for various sectors, including finance, healthcare, and cybersecurity. Through applying LLMs, this section explores the nuanced approaches to detecting such anomalies, ranging from sudden spikes in financial markets to unexpected patient vital signs, providing early warnings of potential issues. LLMs' ability to process and analyze vast amounts of data with temporal dependencies allows for a more sophisticated and accurate detection of anomalies than traditional statistical methods. This exploration covers the technical methodologies employed by LLMs and discusses their implementation challenges and the solutions developed to overcome them. By highlighting the significance of time series anomaly detection, this section aims to provide insights into the advanced capabilities of LLMs, demonstrating their critical role in predictive analytics and their impact on enhancing decision-making processes across various industries.

Dang \textit{et al.} (2020) \cite{dang2020time_series} proposes a BERT based on a natural language processing model to solve the problem of time series anomaly detection; its motivation stems from the similarity between time series anomaly detection and text classification tasks in natural language processing. Simulation results demonstrate that this method only needs a small amount of label data to train the BERT model to obtain better results than the state-of-the-art work.

Dang \textit{et al.} (2021) \cite{dang2021ts-bert_time} introduces the pattern of pre-training and fine-tuning and proposes to adopt the BERT model in the NLP field to model time series, thus addressing the long-distance dependent modeling issue. The performance on two widely used public datasets demonstrates that the method is more accurate on the KPI and Yahoo datasets than the SOTA solutions. This work uses Spectral Residual (SR) to generate labels for unlabeled data. SR is Fourier transform-based and designed for unsupervised anomaly detection in univariate time series. The proposed approach overperforms the SR and SR variation methods regarding the F1 score. The main threat is that this proposed method heavily relies on the pre-training process, which requires massive amounts of data. Therefore, it may not be suitable for anomaly detection task scenarios without much historical data.

\subsection{Anomaly Log Analysis}

This section is dedicated to exploring the capabilities of LLM in scrutinizing log data, a fundamental aspect of maintaining the integrity and performance of IT systems. Log files, generated by various applications, networks, and systems, are rich sources of data that, when analyzed effectively, can unveil operational anomalies, security breaches, and potential system failures. This section discusses how LLMs are revolutionizing the field of log analysis by applying advanced natural language processing techniques to automatically detect, classify, and respond to anomalies within vast, unstructured datasets. The ability of LLMs to understand and interpret the context of log entries enables a more nuanced and efficient anomaly detection process, significantly reducing the time and resources traditionally required for manual log review. By detailing the methodologies, challenges, and success stories of anomaly log analysis using LLMs, this section aims to highlight the transformative impact of these models on cybersecurity, system diagnostics, and operational efficiency, illustrating their indispensable role in modern digital ecosystems.

Chen \textit{et al.} (2022) \cite{chen2022bert-log_anomaly} argues that system logs, which are a primary resource for fault diagnosis and anomaly detection in large-scale computer systems, are challenging to classify due to their unstructured nature. Recent studies have focused on extracting semantic information from these unstructured log messages and converting them into word vectors. However, these methods often overlook the order of words in sequences. To address this, the authors propose BERT-Log, a method that treats the log sequence as a natural language sequence. It uses a pre-trained language model to learn the semantic representation of normal and anomalous logs. A fully connected neural network is then used to fine-tune the BERT model to detect abnormalities. This approach can capture all the semantic information from the log sequence, including context and position. The authors claim that BERT-Log has achieved the highest performance among all the methods on the HDFS dataset, with an F1-score of 99.3\%. They also propose a new log feature extractor on the BGL dataset to obtain log sequences by sliding window, including node ID, window size, and step size. BERT-Log detects anomalies on the BGL dataset with an F1-score of 99.4\%, representing a 19\% performance improvement compared to LogRobust and a 7\% performance improvement compared to HitAnomaly. The model also demonstrated strong generalizability, achieving high F1 scores even when trained on only 1\% of the dataset. The authors conclude that BERT-Log offers better accuracy and generalization ability than previous anomaly detection approaches. They also highlight that their work is the first to utilize node ID and time to form log sequences.

Lee \textit{et al.} (2023) \cite{lee2023lanobert_system} presents a novel approach to log-based anomaly detection using a LLM called LanoBERT\@. The authors argue that existing log-based anomaly detection methods are limited by their inability to capture the complex relationships between log messages and the context in which they occur. LanoBERT addresses this limitation by leveraging the pre-trained BERT model to learn the contextual representations of log messages and their relationships. The authors propose a log sequence representation method that captures the temporal and contextual information of log messages. They also introduce a anomaly detection algorithm that utilizes the learned representations to detect anomalies in log sequences. The authors evaluate LanoBERT on three real-world log datasets. The results show that LanoBERT outperforms existing methods in terms of anomaly detection F1 score and AUROC\@. The authors conclude that LanoBERT is an effective and efficient approach to log-based anomaly detection and has the potential to improve the reliability and security of large-scale computer systems. However, LAnoBERT requires individual training for each log dataset.

Ott \textit{et al.} (2021) \cite{ott2021robust_and} proposes a framework for anomaly detection in log data, aiming to utilize pre-trained general-purpose language models to preserve the semantics of log messages and map them into log vector embeddings. The motivation behind this work is the need for timely and accurate anomaly detection for the reliability, security, safe operation, and mitigation of losses in large computer systems. The challenges include addressing the software evolution due to software upgrades and solving the cold-start problem, where data from the system of interest is unavailable. The rationale for using pre-trained language models is that these representations for the logs are robust and less invariant to log changes, resulting in better generalization of the anomaly detection models. The authors believe that improvements in log vectorization can translate to improvements in the robustness and generalization of the anomaly detection models. The authors present a general framework for learning context and semantic-aware numerical log vector representations suitable for anomaly detection, a comparison of three semantic-level general-purpose language embedding models for anomaly detection, a comparison of two learning objectives for anomaly detection utilizing general language models, a robust model transfer approach for reduction of the false positive rate after software update, and a publicly available implementation of the method and the datasets.

Zhang \textit{et al.} (2023) \cite{zhang2023logprompt_a} proposes a log anomaly detection framework named LogPrompt based on prompt tuning constructs to guide the PLM in learning semantic and sequential information in the logs, improving the evaluation metrics of log anomaly detection tasks. The motivation behind this framework is that traditional log data analysis and anomaly detection are performed manually. Semantic and sequential tokens are comprehensively considered and embedded to help PLM detect point and conditional anomalies effectively and efficiently. Focal loss is used to replace cross-entropy loss, which alleviates the class imbalance of real-world log data. However, methods based on deep learning have significantly progressed, and learning from the labeled log is costly and impractical.

Huang \textit{et al.} (2023) \cite{huang2023improving_log-based} proposes a pre-trained log representation model with hierarchical bidirectional encoder transformers named HilBERT\@. This method parses logs into templates before using the log templates to pre-train HilBERT and design a hierarchical transformer model to capture log template sequence-level information. This work first introduces the design and architecture of the model, which discovers global information and preserves local information. The authors then describe the process of HilBERT pre-training and apply WordPiece tokenization to slice log lines into log sequences. Finally, to utilize HilBERT for anomaly detection tasks, they fine-tune the model with corresponding training data and use the log sequence representation to predict the abnormality of a sequence.

Le \textit{et al.} (2023) \cite{le2023log_parsing} designs appropriate prompts to guide ChatGPT in understanding the log parsing task and extracting the log event/template from the input log messages. In addition, this paper evaluates the effectiveness and the performance of ChatGPT-based log parsing in different scenarios, such as using few-shot scenarios and different prompting methods. The authors designed appropriate prompts to guide ChatGPT in understanding the log parsing task and compared its performance with log parsers in zero-shot and few-shot scenarios.

Gupta \textit{et al.} (2023) \cite{gupta2023learning_representations} introduces BERTOps, an LLM for AI in the operations domain, pre-trained over large-scale public and proprietary log data. The architectural design of BERTOPS is motivated by BERT-BASE\@. The transformer encoder of BERTOps is further trained on log data using the masked language modeling task. After the pretraining of the BERTOps model is complete, it is fine-tuned using a cross-entropy-loss classification for each task.

Shao \textit{et al.} (2023) \cite{shao2022log_anomaly} proposes a Prog-BERT-LSTM model to detect system faults from log text data, improving abnormal logs' detection performance and generalization ability. This approach extracts the log template and then uses the BERT model of the progressive masking strategy to generate the log vectorization representation and combines Mogrifier LSTM with log vector learning sequence features to avoid the loss of sequence features caused by the disappearance of gradient in the calculation process. In the end, Softmax logical regression outputs the predicted abnormal log. This work designs and implements a neural network model combined with dynamic mask ratio and Mogrifier LSTM, which detects log anomalies based on semantic understanding and long-term dependence of sequences. This model uses Magnifier LSTM as the cycle unit, which has the advantages of solid sequence expression ability and simple parameters, and the accuracy is further improved. This paper further enhances the BERT model in log anomaly detection in the Prog-BERT-LSTM model. It introduces the BERT model of progressive masking strategy to vectorize the log sequence to improve the model's training speed and semantic understanding ability.

He \textit{et al.} (2023) \cite{he2023parameter-efficient_log} proposes a new approach for log anomaly detection that efficiently captures semantic information among logs while minimizing training overhead. This approach, named LogBP-LoRA, integrates a pre-training model (BERT) with Low-Rank Adaptation (LoRA) to enhance the detection of anomalies in log data. The goal is to overcome the limitations of traditional BERT models in handling log data and to provide a more resource-efficient solution for anomaly detection in this context. The proposed LogBP-LoRA method aims to solve these challenges by integrating a pre-training model with Low-Rank Adaptation, thereby reducing training overhead while enhancing the model's capability to extract meaningful semantic information from log data. This paper uses a novel LogBP-LoRA approach, which combines a pre-trained BERT model with LoRA specifically for log anomaly detection. The critical innovation is integrating a bypass connection in the self-attention layer of BERT, which allows for efficient training and better semantic information extraction from log data. This method addresses the high computational requirements of traditional BERT models and improves anomaly detection accuracy by effectively capturing the relationships and patterns within log sequences. The approach is validated through extensive experiments on public log datasets, demonstrating its effectiveness and efficiency. However, the proposed method is limited to log data and the complexity of implementing the LogBP-LoRA model in diverse real-world scenarios. Since the model is tailored for log anomaly detection, its applicability might be limited to similar datasets. It may not generalize well to other types of anomaly detection tasks. Additionally, while it addresses the computational efficiency of traditional BERT models, implementing and tuning the proposed method may still present challenges in practical applications, especially in environments with varying log formats and characteristics.

Bobur \textit{et al.} (2020) \cite{bobur2020anomaly_detection} proposed two new methods to detect outliers in a collection of judicial acts and found the second one more fit for the recommender system. The first method for searching for anomalies combines two models: classification and similarity algorithms. The second method shows the usage of the BERT embedding model and the annoy indexing model. This work mentioned that existing model improvements associated with the accuracy and speed of execution for one request, try to use other pre-trained BERT models (token-based sentences), try to change the size of BERT embedding vector, try to use BERT and other similarity distance algorithms. The authors solved the problem of finding anomalies in judicial acts. By comparing different methods, this paper recognized the BERT embedding model is better for their recommender system.

Karlsen \textit{et al.} (2023) \cite{karlsen2023exploring_semantic} applies two feature extraction techniques (syntactic and semantic) in NLP, with no a priori information on the data's log formats. The semantic method (LLM-based Feature Extraction) focuses on extracting meaning from contextual relationships within the log. In contrast, the syntactic approach (TF-IDF-based Feature Extraction) identifies keywords through their frequency in the log file's syntax representation. Semantic LLM comes out on top among the two feature extraction techniques. Semantic extraction yields better results in two out of three datasets and matching performance in the third, albeit at higher computational costs. Future research will aim to reduce this computational cost and refine the sentence embedding representations of log files by exploring and adapting diverse large language models. The lower performances achieved while using the syntactic approach are likely due to the anonymized nature of the dataset, making it more challenging to differentiate abnormal behavior from normal behavior.

Hu \textit{et al.} (2023) \cite{hu2023research_on} explores the intricate challenges of detecting log anomalies in computer systems, an essential task for pinpointing abnormal events or potential issues that may compromise system stability and reliability. This research is propelled by the burgeoning complexity and rapid expansion of software systems, leading to a substantial increase in log data. This surge in data volume has rendered traditional anomaly detection methods inadequate for timely anomaly identification, underscoring the urgency for more refined and universally applicable log anomaly detection techniques. Traditional methods, often reliant on rule-based or statistical approaches, necessitate extensive human oversight, rendering them both time-intensive and less efficacious amidst the deluge of log data from contemporary software systems. Addressing these limitations, the study introduces a pioneering approach, LogADSBERT, which capitalizes on the Sentence-BERT model to distill semantic behavioral attributes from log events, and employs a bidirectional recurrent neural network, specifically Bi-LSTM, for anomaly detection. The paper, however, acknowledges potential validity threats, such as the dependency on the Sentence-BERT model's ability to accurately capture semantic nuances, and the broad applicability of the proposed method across various log data types and software systems. Despite these considerations, experimental outcomes affirm LogADSBERT's superior accuracy over existing log anomaly detection techniques and its resilience in handling novel log event scenarios.

Almodovar \textit{et al.} (2024) \cite{almodovar2024logfit_log} presents LogFiT, an innovative log anomaly detection model that transcends the limitations of traditional methods by eschewing the dependence on log templates and the necessity for labeled data in supervised training. This research is propelled by the critical need for efficient anomaly detection within system logs to uphold the security and reliability of computing systems. Conventional techniques are hampered either by their inability to assimilate semantic information, owing to a reliance on log templates, or by the requirement for extensive labeled datasets for supervised learning, which is often unfeasible. LogFiT is grounded in the concept of utilizing the linguistic insights embedded in a pretrained BERT-based language model, refining it to discern the linguistic patterns characteristic of normal system logs. This methodology enables LogFiT to accommodate the diverse nature of log content and proficiently identify anomalies. By employing a self-supervised learning paradigm, leveraging a pretrained BERT-based framework fine-tuned on patterns of normal log data, LogFiT facilitates effective anomaly detection without necessitating labeled datasets.

Zhang \textit{et al.} (2022) \cite{zhang2022logst_log} proposes a novel representation technique for log semantics employing Sentence-BERT, aimed at enhancing the accuracy of anomaly detection by considering both semantic and word order relationships. This method exhibits consistent performance even with a limited number of labeled normal logs and surpasses previous techniques on the HDFS dataset. The approach seeks to address the challenge of accurately capturing log semantics, a task where existing word embedding methods fall short, by offering a more effective anomaly detection solution in software system logs. The initiative to develop this method stems from the limitations of traditional methods that rely on word embedding and aggregate weighting, which often overlook the semantic relationship dictated by word order and neglect word interactions. The underlying hypothesis is that a more advanced semantic extraction technique will enable a superior understanding and representation of log events. By leveraging Sentence-BERT for semantic representation extraction, the authors aim to maintain the crucial semantic and word order relationships, essential for contextual comprehension of log sequences.

Le \textit{et al.} (2021) \cite{le2021log-based_anomaly} introduces NeuralLog, an innovative technique for anomaly detection in software systems using log analysis. This approach bypasses the conventional requirement for log parsing, aiming to directly derive semantic insights from unprocessed log messages for anomaly detection purposes. The motivation behind NeuralLog stems from the observation that errors in log parsing can significantly undermine anomaly detection performance. By circumventing log parsing, NeuralLog intends to preserve the integrity of log data, thereby enhancing the accuracy of anomaly detection. The challenge that NeuralLog addresses is twofold: first, to effectively interpret the semantic content of raw log messages without the preprocessing step of parsing, and second, to accurately identify anomalies within the vast and complex data environment of software system logs. This task is complicated by the diverse formats and unstructured nature of log messages, which traditionally necessitated parsing to standardize the data for analysis. The rationale for the development of NeuralLog is supported by an empirical study demonstrating the negative impact of log parsing errors on anomaly detection efficacy. By eliminating the log parsing stage, NeuralLog seeks to avoid the potential loss of critical information due to parsing inaccuracies. The approach employs a Transformer-based classification model, capitalizing on its ability to understand contextual relationships in log sequences, which is crucial for identifying anomalies within the logs effectively. The contribution of this work lies in its novel methodology for log-based anomaly detection, which could potentially set a new standard in the field by offering a more reliable and efficient means of identifying system anomalies. This method promises to reduce the time and resources required for anomaly detection by simplifying the preprocessing steps and improving the detection accuracy. However, the study also acknowledges potential threats to its validity, such as the generalizability of the approach across different types of software systems and the effectiveness of the Transformer model in handling the highly variable and domain-specific nature of log data. Further research and extensive testing across varied datasets are necessary to fully evaluate the robustness and applicability of NeuralLog in real-world scenarios.

Huang \textit{et al.} (2020) \cite{huang2020hitanomaly_hierarchical} introduces a hierarchical transformer-based anomaly detection framework designed to analyze system logs by examining both log template sequences and their parameter values. This approach is driven by the critical need for reliable anomaly detection mechanisms within the increasingly complex architectures of contemporary computer systems. Since system log anomalies can significantly affect numerous users and services, developing precise and efficient anomaly detection models is imperative for effective service management and system maintenance. The authors address the limitations of current log-based anomaly detection techniques, which often struggle with unrecognized log templates or overlook the significance of parameter values, resulting in imprecise anomaly identification. The proposed model, HitAnomaly, emerges from the understanding that certain anomalies manifest not only through irregularities in log template sequences but also through unusual parameter values. The authors posit that incorporating the semantic content of log template sequences and the specific parameter values is crucial for identifying a broader spectrum of performance anomalies, suggesting that a model adept at integrating these elements would enhance anomaly detection capabilities. However, the model faces potential challenges, including its ability to generalize across diverse log data types and systems, scale to manage extensive log datasets and adapt to modifications in log formats or system upgrades. Moreover, the model's effectiveness is contingent on the accuracy of the log parser, with parsing inaccuracies posing additional risks to its operational performance.

\subsection{Microservice Anomaly Detection}

In this section, we delve into the sophisticated realm of monitoring and ensuring the reliability of distributed systems through the lens of LLMs. As the architecture of digital services shifts towards microservices—a collection of loosely coupled, independently deployable services—the complexity of detecting anomalies increases significantly. This subsection explores how LLMs are leveraged to navigate this complexity. It offers advanced solutions for identifying discrepancies that may indicate performance issues, failures, or security threats within individual microservices or their interactions. The ability of LLMs to analyze and interpret the vast amounts of data generated by these distributed systems enables a proactive approach to anomaly detection, facilitating early identification and resolution of potential issues. By examining the unique challenges posed by microservice architectures, including dynamic scaling and inter-service communication, this subsection showcases the innovative use of LLMs in enhancing system resilience, security, and operational efficiency. Through detailed case studies and technical insights, this section aims to underscore the transformative impact of LLMs on anomaly detection in modern distributed systems, emphasizing their critical role in maintaining the reliability and security of digital infrastructures.

Sarda \textit{et al.} (2023) \cite{komal2023adarma_auto-detection} proposes a pipeline named ADARMA platform for automatic anomaly detection and remediation based on LLMs aiming to enhance real-time anomaly detection and auto-remediation for microservice deployments. The combination of anomaly detection and auto-remediation reduces downtime and enhances system reliability, resulting in increased productivity and customer satisfaction, which, in turn, drives higher revenue. Prior works have overlooked auto-remediation. The current work focuses on prompt developing and fine-tuning LLMs for auto-remediating anomalies rather than the entire pipeline. In the future, the authors plan to refine detection accuracy, expand remediation tactics, and evaluate the approach's long-term impact.

Khlaisamniang \textit{et al.} (2023) \cite{khlaisamniang2023generative_ai} integrates generative AI technology into self-healing systems, which leverages GPT-4 for automated code generation to enhance the operations of large-scale systems and facilitate automatic repairs. The focus is optimizing system functionality and efficiency at scale while reducing reactive tasks requiring human intervention. The threat might be that ChatGPT was provided with unsuitable prompts to yield unexpected outcomes in log parsing.

\section{Threats}\label{sec:threats}

In the exploration of leveraging LLMs for forecasting and anomaly detection, several significant challenges and deficiencies have become apparent, shaping the landscape of current methodologies and their practical applications. This section delves into the core threats that hinder the effectiveness and reliability of LLMs in these domains. Firstly, the dependence on extensive historical datasets raises concerns about data availability, quality, and the potential for model bias. The issue of generalizability is also critical, as models often struggle to apply learned patterns across diverse contexts or when encountering novel scenarios. Furthermore, the phenomena of hallucination and robustness underscore the models' tendencies to generate misleading or inaccurate outputs under certain conditions, questioning their reliability. The knowledge boundary of LLMs, defined by the scope of their training data, presents another fundamental challenge, limiting their ability to generate insights beyond their informational horizon. Lastly, computational efficiency remains a daunting obstacle, as the resource-intensive nature of these models can restrict their accessibility and scalability. Addressing these threats is paramount for advancing the utility of LLMs in forecasting and anomaly detection, necessitating a multifaceted approach to enhance their performance, reliability, and applicability in real-world settings.

\subsection{Extensive Historical Datasets Dependence}

The dependence on extensive historical datasets stands as a formidable challenge in the deployment of LLMs for forecasting and anomaly detection. This reliance not only necessitates the availability of vast amounts of data but also raises critical issues regarding the representativeness, quality, and bias inherent in the collected information. Historical data, by its nature, may not always encapsulate future trends or rare, anomalous events with sufficient accuracy, leading to models that are potentially myopic or skewed in their predictions and detections. Moreover, the acquisition of such datasets often involves significant financial, legal, and ethical considerations, particularly when dealing with sensitive or proprietary information.

To mitigate these challenges, several strategies can be employed. One approach involves enhancing data diversity and representativeness through techniques such as data augmentation, synthetic data generation, and transfer learning, which can help models generalize better to unseen scenarios. Additionally, employing robust data cleaning and preprocessing methodologies can significantly improve the quality of the datasets, reducing noise and minimizing bias. Active learning and few-shot learning techniques offer promising avenues to reduce the dependency on large datasets by enabling models to learn effectively from smaller, more targeted data samples. Lastly, the development of models that can dynamically update and incorporate new data streams can help alleviate the reliance on static historical datasets, making them more adaptive to evolving trends and patterns.

Addressing the extensive historical datasets dependence not only involves technical and methodological advancements but also a concerted effort to ensure ethical data practices, emphasizing transparency, fairness, and inclusivity in data collection and model training processes. By tackling these issues head-on, the field can move towards more reliable, efficient, and equitable forecasting and anomaly detection solutions that are less tethered to the limitations of their underlying data.

\subsection{Generalizability}

Generalizability emerges as a pivotal concern in harnessing LLMs for forecasting and anomaly detection, highlighting the challenge of applying insights derived from specific datasets across varied contexts and domains. This issue is particularly pronounced when models trained on data from one domain or time period are expected to perform accurately on data from another, often leading to suboptimal predictions and detections. The root of this challenge lies in the models' ability to abstract and transfer learned patterns to new, unseen environments, a task that is not trivial given the complex and dynamic nature of real-world data.

To enhance the generalizability of LLMs, several strategies can be considered. Developing models with a stronger emphasis on domain adaptation techniques allows for more flexible adjustments to different data characteristics, potentially improving performance across diverse settings. Incorporating multi-task learning frameworks can also aid in this endeavor by enabling models to learn from a variety of tasks simultaneously, fostering a broader understanding that can be applied to new problems. Further, the application of meta-learning approaches, where models learn to learn, offers a pathway to quickly adapt to new domains with minimal data requirements. Another solution lies in the rigorous evaluation of models across heterogeneous datasets and conditions prior to deployment, ensuring their robustness and adaptability.

Investing in these approaches not only addresses the immediate challenge of generalizability but also contributes to the development of more versatile and resilient forecasting and anomaly detection systems. By prioritizing the creation of models that can navigate the nuances of different domains with greater ease, researchers and practitioners can expand the applicability and effectiveness of LLMs, paving the way for innovations that are both impactful and enduring across a multitude of scenarios.

\subsection{Hallucination and Robustness}

The phenomena of hallucination and robustness in LLMs for forecasting and anomaly detection underscore a critical vulnerability: the tendency of these models to generate false or misleading information (hallucinations) and their susceptibility to performance degradation under adversarial or noisy conditions. Hallucination challenges the credibility of model outputs, as LLMs might produce plausible yet entirely fabricated data points or trends, leading to misguided decisions or analyses. Similarly, a lack of robustness signifies that minor perturbations in the input data or adversarial attacks could significantly impair the model's accuracy and reliability, jeopardizing its utility in sensitive or critical applications.

Addressing these issues requires a multifaceted approach focused on enhancing the integrity and resilience of model outputs. Implementing rigorous validation and verification mechanisms can help in identifying and mitigating hallucinations, ensuring that model predictions are grounded in the data. Techniques such as adversarial training, where models are exposed to and learn from perturbed or challenging inputs during training, can improve robustness by preparing the model for a wider array of input scenarios. Furthermore, incorporating uncertainty quantification methods allows for a better assessment of the confidence in model outputs, providing users with valuable context regarding the reliability of predictions and detections.

The development of interpretability tools and frameworks also plays a crucial role, as understanding the reasoning behind model outputs can help in diagnosing and correcting for hallucinations and vulnerabilities. By investing in these strategies, the field can advance towards creating LLMs that not only excel in forecasting and anomaly detection tasks but do so with a higher degree of trustworthiness and resilience, marking a significant step forward in the practical deployment of these technologies.

\subsection{Knowledge Boundary}

The concept of the knowledge boundary in the context of LLMs for forecasting and anomaly detection refers to the inherent limitations of these models to generate insights or predictions beyond the scope of their training data. This limitation poses a significant challenge, as LLMs may struggle to accurately address novel events, emerging trends, or previously unseen anomalies, leading to potential gaps in their predictive capabilities. The knowledge boundary essentially demarcates the frontier of the model's understanding, beyond which its reliability and accuracy can sharply decline. This is particularly problematic in rapidly evolving domains or in situations where the future does not neatly reflect the past.

To extend the knowledge boundary of LLMs, several strategies can be implemented. One approach is continuous or incremental learning, where models are routinely updated with new data, allowing them to adapt to recent developments and incorporate emerging patterns into their predictions. Another strategy involves leveraging transfer learning, where a model trained on one task is adapted for another, potentially related task, thereby utilizing its pre-existing knowledge base to bridge gaps in understanding. Additionally, employing ensemble methods that combine the outputs of multiple models can help in mitigating the knowledge boundary issue, as different models may capture varied aspects of the data, providing a more comprehensive overview.

The integration of external knowledge bases or expert systems with LLMs offers another promising solution, where models can access and incorporate specialized knowledge that may not be present in their training datasets. Furthermore, developing models with advanced reasoning capabilities and the ability to query external sources when faced with unknowns can enhance their ability to navigate beyond their initial knowledge boundaries.

By adopting these strategies, the field can make strides towards developing LLMs with broader, more flexible knowledge bases, significantly enhancing their utility and effectiveness in forecasting and anomaly detection across a wider range of scenarios and domains.

\subsection{Computational Efficiency}

The challenge of computational efficiency in deploying LLMs for forecasting and anomaly detection cannot be overstated. The sheer scale and complexity of these models demand substantial computational resources, making them less accessible for many organizations and potentially limiting their scalability and practicality for real-time applications. High computational costs are associated not only with training these models but also with their inference, especially when processing large volumes of data or requiring rapid response times. This computational burden poses significant hurdles, particularly for small to medium-sized enterprises or in scenarios where computational resources are constrained.

Addressing the computational efficiency of LLMs involves a multi-pronged approach. Model optimization techniques, such as pruning, quantization, and knowledge distillation, can significantly reduce model size and complexity while maintaining performance, making the models lighter and faster for both training and inference phases. Additionally, adopting efficient architectures specifically designed for speed and low resource consumption, such as transformer variants optimized for efficiency, can further alleviate computational demands.

Leveraging hardware acceleration through the use of GPUs, TPUs, and specialized inference chips offers another avenue to enhance computational efficiency. These technologies can dramatically speed up model computations, making it feasible to deploy LLMs in more resource-sensitive environments. Furthermore, the development of cloud-based solutions and edge computing allows for the distribution of computational tasks, optimizing resource usage across networks and devices, thereby reducing the overall computational load on individual systems.

Efforts to improve algorithmic efficiency, through advancements in model design and training methodologies, also play a critical role. Techniques that enable more data-efficient learning, such as few-shot learning or transfer learning, can reduce the need for extensive computation by minimizing the amount of data required to train or adapt models effectively.

By focusing on these strategies, the research and development community can make significant strides towards creating LLMs that are not only powerful and accurate but also computationally efficient, ensuring their wider accessibility and applicability in a diverse range of forecasting and anomaly detection tasks.

\section{Future Directions and Trends}\label{sec:future}

As the field of LLMs continues to evolve, its application in forecasting and anomaly detection is poised for transformative advancements. The convergence of technological innovation, research breakthroughs, and interdisciplinary collaboration heralds a future where LLMs can offer unprecedented accuracy, adaptability, and insight. This section outlines key future directions and trends that are expected to shape the utilization of LLMs in these domains.

\bulletparagraph{Integration of Multimodal Data Sources}

The future of LLMs in forecasting and anomaly detection is likely to see a significant shift towards the integration of multimodal data sources. By combining textual data with visual, auditory, and sensor-based information, LLMs can develop a more holistic understanding of complex phenomena. This multimodal approach could enhance the models' ability to detect nuanced anomalies and forecast events with greater precision, leveraging the complementary strengths of diverse data types, and benefit data automated validation process~\cite{zhang2023automated}.

\bulletparagraph{Advancements in Transfer and Meta-Learning}

Transfer and meta-learning represent promising avenues for making LLMs more adaptable and efficient. Future developments in these areas could enable models to swiftly adjust to new domains or tasks with minimal additional training. Such capabilities would be invaluable in rapidly changing environments or in applications where data scarcity poses a challenge \cite{zhang2023making, sun2016migrating}. By improving the versatility of LLMs, these techniques can expand their applicability across a wider range of forecasting and anomaly detection scenarios.

\bulletparagraph{Focus on Explainability and Trustworthiness}

As LLMs assume a more prominent role in decision-making processes, the demand for explainability and trustworthiness will intensify. Future research is expected to prioritize the development of models that not only perform with high accuracy but also provide transparent and interpretable explanations for their outputs. Enhancing the explainability of LLMs can build trust among users \cite{10.1609/aaai.v37i5.25748}, facilitate the identification of biases, and ensure the ethical application of these technologies.

\bulletparagraph{Medical Analysis}

Recent research in medical analysis has achieved notable advancements in image segmentation, classification, and trend prediction through end-to-end applications of deep learning and machine learning techniques \cite{dong2021variational, zhang2024deepgi}. Medical imaging data, including Computed Tomography (CT), Magnetic Resonance Imaging (MRI), and Optical Coherence Tomography (OCT) \cite{lin2022deep}, often consist of multilayer scan results, with pathological changes potentially distorting these results. Accurately segmenting or classifying such lesions necessitates extensive training on a large corpus of manually annotated medical images, enabling the model to learn pathological changes end-to-end. Interestingly, the task of identifying pathologies in medical images bears a strong resemblance to abnormality detection utilizing LLMs. This parallel raises a compelling research challenge: how can prior knowledge embedded in LLMs be harnessed to enhance learning efficiency, particularly in the context of limited labeled medical imagery?

Concurrently, advancements in medical imaging technology have significantly enhanced imaging quality and increased the volume of data available \cite{chen2021high, chen2022high, chen2023ultrahigh}. Despite these advances, efforts to integrate medical data with LLMs have primarily focused on individual images \cite{PRIOR2023_Cheng, 10292600}. The task of employing LLMs to process and analyze large-scale medical image volumes, which often include multilayer information, continues to pose a significant challenge. Future research directions could explore the development of LLMs specifically tailored to navigate and interpret the complexities introduced by these technological advancements in medical imaging.

As LLMs holds the potential to revolutionize doctor-patient interactions, paving the way for more efficient and comprehensive communication channels. With the assistance of LLMs, healthcare professionals can anticipate improved patient education, enhanced clarity in medical discussions, and streamlined dissemination of complex medical information.

\bulletparagraph{Real-time Processing and Edge Computing}

The ability to process data in real-time and deploy LLMs closer to data sources, such as through edge computing, is set to become a crucial trend. This shift towards real-time analysis and decentralized processing can significantly reduce latency, increase the timeliness of insights, and enable the deployment of LLMs in environments where immediate responses are critical. It also opens the door to new applications in sectors such as finance, healthcare, and manufacturing, where quick decision-making is paramount \cite{su2023edgegym}. As a concrete example, in the realm of indoor positioning, LLMs have the potential to discern patterns in edge data, such as trends in WiFi signals, thereby augmenting the precision of existing WiFi signal-based indoor positioning systems \cite{lin2018indoor}. This capability demonstrates the extensive applicability of LLMs in leveraging real-time edge data for improved accuracy in critical applications.

\bulletparagraph{Sustainable and Energy-efficient Modeling}

As computational demands continue to grow, the sustainability and energy efficiency of LLMs will become a pressing concern. Future trends are likely to include the pursuit of more environmentally friendly models through optimized algorithms, energy-efficient hardware, and practices that minimize the carbon footprint of training and deploying LLMs. This focus on sustainability is essential for ensuring that the benefits of LLMs can be realized without exacerbating environmental impacts.

In conclusion, the trajectory of LLMs in forecasting and anomaly detection is marked by exciting opportunities and challenges. By embracing these future directions and trends, the field can unlock new potentials, addressing pressing issues while paving the way for innovative applications that leverage the full capabilities of LLMs in understanding and predicting complex systems.

\bulletparagraph{Computer vision}

LLMs have made significant strides in aligning with images and performing tasks such as classification and semantic segmentation \cite{radford2021learning, kirillov2023segany, liao4583223self}. However, within the field of computer vision, several low-level visual processing tasks persist, which pose challenges in establishing direct relationships with semantics, thus hindering the direct application of LLMs to these tasks.

These low-level visual tasks encompass various aspects, including denoising, defect detection, multi-view reconstruction, and more \cite{liu2023unveiling, shangguan2020dog}. Despite their crucial role in image processing and analysis, these tasks often involve intricate visual patterns and features that are not easily discernible through high-level semantic representations alone. As a result, incorporating LLMs into these tasks remains challenging due to the inherent disparity between low-level visual processing and semantic understanding.

\bulletparagraph{Collaboration Across Disciplines}

The future development and application of LLMs in forecasting and anomaly detection will benefit greatly from increased collaboration across different fields, including multi-core system \cite{zhang2015approximate}, statistics, machinery \cite{zang2024precision, herickhoff2019low}, robotics~\cite{liu2024particle}, other domain-specific areas \cite{chiang2023two, ni2024smartfix, qiao2023application, popokh2021illumicore, wei2024strategic, hu2023m, sun2014effects}, and ethics \cite{ijcai2023p337, zang2024evaluating, zheng2021makes}. Such interdisciplinary efforts can enrich the models with diverse perspectives and expertise, leading to more robust, innovative, and ethically sound solutions.

\section{Related Surveys and Reviews}\label{sec:related}

With the rapid advancement of LLMs, considerable comprehensive reviews have appeared, offering in-depth analyses of different facets of this technology. Zhao \textit{et al.} \cite{zhao2023a_survey} provide an extensive overview of LLMs, detailing their background, fundamental discoveries, and core technologies, summarizing a broad spectrum of existing research. Conversely, Yang \textit{et al.} \cite{yang2023harnessing_the} concentrate on the application spectrum of LLMs across various downstream tasks, highlighting the deployment challenges that accompany their use. Chang \textit{et al.} \cite{chang2024a_survey} focus on the evaluation methodologies for LLMs, exploring the criteria, contexts, and methodologies for assessing their performance in downstream applications and societal impacts. Chang and Bergen \cite{chang2023language_model} delve into the abilities and constraints of LLMs across varied downstream tasks. Huang \textit{et al.} \cite{huang2023towards_reasoning} review the progress in enhancing and assessing the reasoning capabilities of LLMs. These studies collectively address multiple aspects of LLMs, like training, evaluation, and application to different domains. However, before this paper, the burgeoning and promising domain of LLM-based Agents had not been focused specifically. This work compiles over 40 latest relevant works on LLM-based forecaster and anomaly detectors, encapsulating their development, applications, and evaluation processes.

\section{Conclusion}\label{sec:conclusion}

This systematic literature review has explored the burgeoning field of LLMs in the context of forecasting and anomaly detection, offering a comprehensive overview of current methodologies, challenges, and future directions. As we have seen, LLMs hold immense potential for transforming these domains, providing sophisticated tools capable of parsing vast datasets to predict future events and identify deviations from norms with remarkable accuracy. However, the journey is fraught with challenges, including the dependence on extensive historical datasets, issues of generalizability, the occurrence of hallucinations, knowledge boundaries, and the need for computational efficiency.

Despite these obstacles, the path forward is illuminated by promising solutions and innovations. The integration of multimodal data sources, advancements in transfer and meta-learning, a focus on explainability and trustworthiness, the push towards real-time processing and edge computing, interdisciplinary collaboration, and a commitment to sustainable modeling practices all represent key trends that will shape the future of LLMs in forecasting and anomaly detection.

The review underscores the importance of continued research and development in this area, highlighting the need for models that are not only powerful and accurate but also transparent, adaptable, and accessible. As technology advances, so too must our approaches to ethical considerations, ensuring that the deployment of LLMs contributes positively to society and does not exacerbate existing inequalities or environmental issues.

In conclusion, the potential of LLMs to revolutionize forecasting and anomaly detection is clear, yet realizing this potential requires a concerted effort across the scientific community, industry stakeholders, and policymakers. By addressing the challenges outlined in this review and harnessing the opportunities presented by emerging trends, we can look forward to a future where LLMs play a pivotal role in navigating the complexities of the modern world, driving insights and innovations that benefit all of society.

\bibliographystyle{unsrt}

\end{document}